\documentclass[twoside]{article}

\usepackage[accepted]{aistats2022}
\usepackage[utf8]{inputenc} 
\usepackage[T1]{fontenc}    
\usepackage{hyperref}       
\usepackage{url}            
\usepackage{booktabs}       
\usepackage{amsfonts}       
\usepackage{nicefrac}       
\usepackage{microtype}      
\usepackage{xcolor}         

\usepackage{microtype}
\usepackage{graphicx}
\usepackage{booktabs} 

\usepackage{microtype}
\usepackage{graphicx}
\usepackage{booktabs} 

\usepackage{natbib}
\usepackage{amssymb, amsmath,amsthm}
\usepackage{amsfonts}
\usepackage{algorithm}
\usepackage{algorithmic}
\usepackage{paralist}
\usepackage{multirow}
\usepackage{times}
\usepackage{wrapfig}

\usepackage{url,enumerate}
\usepackage{color,xcolor}
\usepackage{makeidx}  
\usepackage{amsmath,amssymb}
\usepackage{mathtools}
\usepackage[small, compact]{titlesec}
\usepackage{xspace}
\usepackage{epstopdf}
\usepackage{cite}
\usepackage{mathrsfs}
\usepackage{times}
\usepackage{enumerate}
\usepackage{color}
\usepackage{graphicx,epsfig}
\usepackage{amsmath,amssymb,xspace}
\usepackage{url}
\usepackage{hyperref}
\usepackage{bm}
\usepackage{bbm}
\usepackage{upgreek}
\usepackage{cleveref}
\usepackage{multirow}
\usepackage{ulem}
\usepackage{cancel}
\usepackage{subcaption}
\usepackage{dsfont}
\usepackage{hyperref}

\newcommand{\ours}{{DMFAL}\xspace}

\newcommand{\cmt}[1]{}
\newcommand{\eg}{{\textit{e.g.},}\xspace}
\newcommand{\ie}{{\textit{i.e.},}\xspace}
\newcommand{\etc}{{\textit{etc}.}\xspace}

\newcommand{\diag}{{\rm diag}}

\renewcommand{\vec}{{\rm vec}}

\newcommand{\bepsilon}{\boldsymbol{\epsilon}}
\newcommand{\boldeta}{\boldsymbol{\eta}}
\newcommand{\brho}{\boldsymbol{\rho}}

\renewcommand{\c}{{\bf c}}

\newcommand{\h}{{\bf h}}

\newcommand{\s}{{\bf s}}

\renewcommand{\u}{{\bf u}}

\newcommand{\x}{{\bf x}}
\newcommand{\y}{{\bf y}}

\newcommand{\A}{{\bf A}}

\newcommand{\Dcal}{\mathcal{D}}

\newcommand{\I}{{\bf I}}
\newcommand{\J}{{\bf J}}

\renewcommand{\L}{{\bf L}}
\newcommand{\Lcal}{{\mathcal{L}}}

\newcommand{\Ocal}{{\mathcal{O}}}

\newcommand{\N}{\mathcal{N}}  

\renewcommand{\S}{{\bf S}}

\newcommand{\barf}{{\widehat{\h}}}

\newcommand{\barV}{{\widehat{\V}}}
\newcommand{\barJ}{{\widehat{\J}}}

\newcommand{\bary}{{\widehat{\y}}}
\newcommand{\baralpha}{{\widehat{\balpha}}}
\newcommand{\barA}{{\widehat{\A}}}
\newcommand{\bareps}{{\widehat{\bepsilon}}}
\newcommand{\V}{{\bf V}}
\newcommand{\W}{{\bf W}}
\newcommand{\Wcal}{{\mathcal{W}}}

\newcommand{\Xcal}{{\mathcal{X}}}

\newcommand{\Ycal}{{\mathcal{Y}}}

\newcommand{\balpha}{\boldsymbol{\alpha}}

\newcommand{\btheta}{\boldsymbol{\theta}}

\newcommand{\bxi}{\boldsymbol{\xi}}
\newcommand{\bSigma}{\boldsymbol{\Sigma}}

\newcommand{\bOmega}{\boldsymbol{\Omega}}

\newcommand{\bmu}{\boldsymbol{\mu}}

\newcommand{\0}{{\bf 0}}

\newcommand{\ben}{\begin{enumerate}}
\newcommand{\een}{\end{enumerate}}

\newcommand{\argmax}{\operatornamewithlimits{argmax}}

\newcommand{\EE}{\mathbb{E}}

\begin{document}
	
\twocolumn[

\aistatstitle{Deep Multi-Fidelity Active Learning of High-Dimensional Outputs}

\aistatsauthor{ Shibo Li \And Robert M. Kirby \And Shandian Zhe }

\aistatsaddress{ School of Computing, University of Utah \\
\texttt{ \{shibo, kirby, zhe\}@cs.utah.edu}  }

 ]

\begin{abstract}
	Many applications, such as in physical simulation and engineering design,  demand we estimate functions with high-dimensional outputs. To reduce the expensive cost of generating training examples, we usually choose several fidelities to enable a cost/quality trade-off. In this paper, we consider the active learning task to  automatically identify the fidelities and training inputs to query new example so as to achieve the best learning benefit-cost ratio. To this end, we propose \ours, a \underline{D}eep \underline{M}ulti-\underline{F}idelity \underline{A}ctive \underline{L}earning approach. We first develop a deep neural network based multi-fidelity model for high-dimensional outputs, which can flexibly capture strong complex correlations across the outputs and  fidelities to enhance the learning of the target function. We then propose a mutual information based acquisition function that extends the predictive entropy principle. To overcome the computational challenges caused by large output dimensions, we use the multi-variate delta method and moment-matching to estimate the output posterior, and Weinstein-Aronszajn identity to calculate and optimize the acquisition function. The computation is tractable and efficient. We show the advantage of our method in several applications of computational physics and engineering design. 
\end{abstract}
\section{Introduction}
Many applications require us to compute a mapping from low-dimensional inputs to high-dimensional outputs. For example, topology optimization~\citep{rozvany2009critical} aims to find an optimal structure (high-dimensional output) given several design parameters (low-dimensional input). Physical simulation uses numerical solvers to solve partial differential equations (PDEs), which maps the PDE parameters and parameterized initial/boundary conditions (low dimensional input) to the high-dimensional solution field on a mesh. The exact computation of these mappings is often costly. Hence,  learning a surrogate model to directly predict the output (rather than computing from scratch every time) is of great interest and importance~\citep{kennedy2000predicting,conti2010bayesian}.

However, collecting training examples is a bottleneck, because to obtain these examples we still have to  conduct the original, expensive computation (\eg running numerical solvers). To reduce the cost, we can compute the training examples at different fidelities to enable a trade-off between the cost and quality. Low fidelity examples are cheap to acquire but inaccurate while high-fidelity examples are much more accurate yet expensive. Despite the disparity in accuracy, the outputs at different fidelities are strongly correlated. Hence, examples from different fidelities can all be useful in learning the target mapping. 


To reduce the cost while maximizing the learning performance, we develop \ours, a deep multi-fidelity active learning approach that  can identify both the fidelity and input location to query (or generate) new training examples so as to achieve the best benefit-cost ratio. To our knowledge, this is the first work that incorporates multi-fidelity queries in active learning of high-dimensional outputs (See Fig. 1 in the Appendix for more illustrations). Our work naturally extends the standard active learning, where the query cost is assumed uniform, and minimizing the total cost is equivalent to minimizing the number of queries. 

Specifically,  we first propose an expressive deep multi-fidelity model. We use a chain of neural networks (NNs) to model the outputs in each fidelity. Each NN  generates a low-dimensional latent output first, and then projects it to the high-dimensional observational space. Both the original input and latent output are  fed into the NN of the next fidelity so that we can efficiently propagate information throughout the fidelities and flexibly capture their complex relationships. Second, we propose an acquisition function based on the mutual information of the outputs between each fidelity and the highest fidelity (at which we predict the target function). This can be viewed as  an extension of the predictive uncertainty principle for the traditional, singe-fidelity active learning. When we seek to query with the highest fidelity, the acquisition function is reduced to the output entropy. We found empirically our acquisition function outperforms the popular BALD~\citep{houlsby2011bayesian} and predictive variance principle~\citep{gal2017deep} adapted to the multi-fidelity setting.  Third, we address the challenges of computing and optimizing the acquisition function. Due to the large output dimension, it is very expensive or even infeasible to estimate the required covariance and cross covariance matrices with popular Monte-Carlo (MC) Dropout~\citep{gal2016dropout} samples. To overcome this problem, we consider the NN weights in the latent output layers as random variables and all the other weights as hyper-parameters. We develop a stochastic structural variational learning algorithm to jointly estimate the hyper-parameters and posterior of the random weights. We then use the multi-variate delta method to compute the moments of the hidden outputs in each fidelity, and use moment-matching to estimate a joint Gaussian posterior of the hidden outputs. We use  Weinstein-Aronszajn identity to compute the entropy of their projection --- the observed high-dimensional outputs. In this way, we can analytically calculate and optimize the acquisition function in a tractable, reliable and efficient way.  

For evaluation, we examined \ours in three benchmark tasks of computational physics, a topology structure optimization problem, and a computational dynamic fluids (CFD) application to predict flow velocity fields.\cmt{ of flows. driven by rectangular boundaries.} The output dimensions of these applications vary from hundreds to hundreds of thousands. Our method consistently achieves much better learning performance with the same query cost, as compared with using random query strategies, approximating  the acquisition function with dropout samples of the latent outputs, and using other acquisition functions. Our learned surrogates gain 40x and 460x speed-up in solving  (predicting) optimal structures and velocity fields than running standard numerical methods.  Even counting the total cost of active training, the per-solution cost of \ours rapidly becomes far below that of the numerical methods with the increase of acquired solutions. 

\section{Background}
\textbf{Problem setting.}  We assume that we can query training examples with $M$ fidelities, which correspond to $M$ mappings, $\{\y_m(\x) \in \mathds{R}^{d_m}\}_{1 \le m \le M}$ where $\x \in \Xcal$ is an $r$-dimensional input, $r$ is small, $d_m$ is the output dimension, often very large, and $\{d_m\}$ are not necessarily identical.  In general, we assume $d_1 \le \ldots \le d_M$.  For instance, in structure design, the input can be  the design parameters. The high fidelity examples correspond to high-resolution structures while the low fidelity ones low-resolution structures.  We denote  by $\lambda_m$ the cost to query with fidelity $m$. Our goal is to estimate $\y_M(\x)$ (\ie the target function). Among the examples at different fidelities are strong (but complex) correlations, due to the underlying common bases, say, in physics. Hence, all these examples are valuable to estimate the target function.  To perform active learning, we begin with a small set of examples with mixed fidelities. Each step, we select both the fidelity and input location to query (or generate) new examples, so as to best balance the learning improvement (on $\y_M(\x)$) and computational cost, \ie maximizing the benefit-cost ratio. 

\textbf{Dropout active learning.} A successful application of active learning with deep neural networks is image classification~\citep{gal2017deep,kirsch2019batchbald}. Typically, a pool of unlabeled input examples, \ie images,  is pre-collected. Each time, we rank the input examples according to an acquisition function computed based on the current model. We then query the labels of the top examples and add them into the training set. There are two popular acquisition functions. The first one is the predictive entropy of the label, 
\begin{align}
	\mathbb{H}[y|\x, \Dcal] = -\sum_{c} p(y=c|\x, \Dcal) \log p(y=c|\x, \Dcal), \label{eq:image-ac-1}
\end{align}
where $\x$ is the input, and $\Dcal$ are the training data. The other one is BALD (Bayesian active learning by disagreement)~\citep{houlsby2011bayesian}, namely the mutual information between the label and model parameters, 
\begin{align}
	\mathbb{I}[y, \btheta|\x, \Dcal]  = \mathbb{H}[y|\x, \Dcal]- \EE_{p(\btheta|\Dcal)}\left[\mathbb{H}(\y|\x, \btheta, \Dcal)\right], \label{eq:image-ac-2}
\end{align}
where $\btheta$ are the NN weights. To calculate these acquisition functions, we need to first estimate the posterior of the NN output and weights. A popular approach is to use Monte-Carlo (MC) Dropout~\citep{gal2016dropout}, which is essentially a variational inference method of Bayesian NNs. MC Dropout uses a probability $p_i$ to randomly drop the neurons in each layer $i$ (equivalent to zeroing out the corresponding weights) in the forward pass and performs backward pass over the remaining neurons. After training, the weights and output obtained from one such forward pass can be viewed as a sample of the approximate posterior. We can run Dropout multiple times to collect a set of posterior samples and calculate the acquisition functions by Monte-Carlo approximations. 

\vspace{-0.05in}
\section{Deep Multi-Fidelity Modeling for High-Dimensional Outputs}
\vspace{-0.05in}
Despite its success, dropout active learning might be inappropriate for high-dimensional (continuous) outputs. When the output is a continuous vector, it is natural to use a multivariate Gaussian posterior, $p(\y|\x, \Dcal) \approx \N(\y|\bmu(\x), \bSigma(\x))$\footnote{we can also use a multivariate student $t$ distribution, but the problem remains. }, and the entropy term $\mathbb{H}(\y|\x, \Dcal)$ in the acquisition functions  (see \eqref{eq:image-ac-1} and \eqref{eq:image-ac-2}) require us to compute the log determinant of the covariance matrix, $\log|\bSigma(\x)|$. While we can use dropout samples to construct an empirical  estimate $\widehat{\bSigma}(\x)$, due to the large output dimension $n$, the computation of this $n\times n$ matrix and its log determinant is very expensive  or even infeasible. 
One might seek to only estimate the predictive variance of each individual output. However, this will ignore the strong output correlations, which is critical to effectively evaluate the uncertainty and calculate the acquisition function. Furthermore, in our active learning task, we want to optimize the acquisition function to find the best input (rather than rank the inputs in a pre-collected pool). The empirical covariance estimate from random samples can lead to numerical instability in optimization and inconsistent results from different runs. 

To overcome these problems for multi-fidelity active learning, we first propose an expressive deep multi-fidelity model, for which we develop a structural variational inference method to capture the posterior dependency of the outputs. We propose a novel acquisition function to allow multi-fidelity queries, and develop an efficient and reliable approach to calculate the acquisition function. 

Specifically, we use deep neural networks to build a multi-fidelity high-dimensional output model (see the graphical representation in Fig. 2 of the Appendix) that can flexibly, efficiently capture complex relationships between the outputs and between the fidelities, taking advantage of these relationships to enhance the learning at the highest fidelity (\ie target function).  For each fidelity $m$, we introduce a neural network (NN) that first generates a  $k_m$ dimensional  latent output, and then projects it to the $d_m$ dimensional observational space, where $k_m \ll d_m$. The NN is parameterized by $\{\W_m, \btheta_m, \A_m\}$ where $\W_m$ is a $k_m \times l_m$ weight matrix in the latent output layer, $\A_m$ a $d_m \times k_m$ projection matrix, and $\btheta_m$ the weights in all the other layers. Denote by $\bxi_m$ the NN input, by $\h_m(\x)$ the latent NN output, and by $\y_m(\x)$ the observed output. The model is defined by
\begin{align}
&\bxi_m = [\x; \h_{m-1}(\x)], \;\; \h_m(\x) = \W_m\brho_{\btheta_m}(\bxi_m), \notag \\
&\y_m(\x) = \A_m \h_m(\x) + \bepsilon_m, \label{eq:our-model}
\end{align}
where $\bxi_1 = \x$,  $\brho_{\btheta_m}$  is the second last layer of the NN at fidelity $m$, of dimension $l_m$, and $\bepsilon_m \sim \N(\bepsilon_m|\0, \sigma^2_m\I)$ is an isotropic Gaussian noise. Note that $\brho_{\btheta_m}$ can be considered as $l_m$  nonlinear basis functions  parameterized by $\btheta_m$. Through their combinations via $\W_m$ and $\A_m$, we can flexibly capture  the complex relationships between the elements of $\y_m$ to improve the prediction. Furthermore,  the input $\bxi_{m}$ is the concatenation of the original input $\x$ and the latent output $\h_{m-1}$. The latter can be viewed as a compact  (or low-rank) summary of all the information up to fidelity $m-1$.  Through an NN mapping, we obtain the latent output $\h_m = \text{NN}(\x, \h_{m-1}(\x))$ --- the summary up to fidelity $m$ --- and then generate the high dimensional output $\y_m$. Thereby, we are able to efficiently integrate the information from lower fidelities and grasp the strong, complex relationship between the current and prior fidelities.

We place a standard Gaussian prior over the elements in each $\W_m$. Similar to~\citep{snoek2015scalable}, we consider all the remaining parameters as hyper-parameters to ease the posterior inference and uncertainty reasoning. Given the training dataset $\Dcal = \{\{(\x_{nm}, \y_{nm})\}_{n=1}^{N_m}\}_{m=1}^M$, the joint probability  of our model is given by
\begin{align}
	&p(\Wcal, \Ycal|\Xcal, \Theta, \s) = \prod\nolimits_{m=1}^M \N\left(\vec(\W_m)|\0, \I\right) \notag  \\
	 &\cdot \prod\nolimits_{n=1}^{N_m} \N\big(\y_{nm}|\A_m \h_m(\x_{nm}), \sigma_m^2\I\big), \label{eq:joint-prob}
\end{align}
where $\Wcal = \{\W_m\}_{m=1}^M$,  $\Theta=\{\btheta_m, \A_m\}_{m=1}^M$, $\s = [\sigma^2_1, \ldots, \sigma^2_M]$, and $\{\Xcal$, $\Ycal\}$ are the inputs and outputs in $\Dcal$. 
To estimate the posterior of our model (which is used to calculate the acquisition function and query new examples), we develop a stochastic structural variational learning algorithm. Specifically, for each $\W_m$,  we introduce a multivariate Gaussian variational posterior, $q(\W_m) = \N\left(\vec(\W_m)|\bmu_m, \bSigma_m\right)$.  To ensure the positive definiteness, we parameterize  $\bSigma_m$ by its Cholesky decomposition, $\bSigma_m = \L_m \L_m^\top$, where $\L_m$ is a lower triangular matrix. We then assume the posterior $q(\Wcal) =\prod_{m=1}^M q(\W_m)$, and construct a variational model evidence lower bound (ELBO)~\citep{wainwright2008graphical}, $\Lcal\big(q(\Wcal), \Theta, \s\big) = \EE_{q}\left[ \log{p(\Wcal, \Ycal|\Xcal, \Theta, \s)}/{q(\Wcal)}\right] =-\text{KL}\big(q(\Wcal)\|p(\Wcal)\big) + \sum_{m=1}^M\sum_{n=1}^{N_m} \EE_q\left[\log \N(\y_{nm}|\A_m \h_m(\x_{nm}),\sigma_m^2\I)\right]$, where $\text{KL}(\cdot \| \cdot)$ is Kullback Leibler divergence and $p(\Wcal)$ the prior of $\Wcal$. We maximize $\Lcal$ to jointly estimate $q(\Wcal)$ and the hyper-parameters. While $\Lcal$ is intractable, we use the reparameterization trick~\citep{kingma2013auto} to generate parameterized samples for each $\vec(\W_m)$: $\bmu_m + \L_m \boldeta_m$ where $\boldeta_m \sim \N(\0, \I)$ to obtain a stochastic estimate of $\Lcal$,  and conduct stochastic optimization. 
 
\section{Multi-Fidelity Active Learning}
\subsection{Mutual Information based Acquisition Function}
We now consider how to perform active learning with multi-fidelity queries. We assume that at each fidelity $m$, the most valuable training example is the one that can best improve our prediction of the target function, namely, the prediction at the highest fidelity $M$. 
To this end, we propose our acquisition function based on the mutual information between the outputs at fidelity $m$ and $M$,
\begin{align}
	&a(\x, m)  = \frac{1}{\lambda_m} \mathbb{I}\left(\y_m(\x),  \y_M(\x)|\Dcal\right) \notag \\
	&= \frac{1}{\lambda_m} \big(\mathbb{H}(\y_m|\Dcal) +\mathbb{H}(\y_M|\Dcal) - \mathbb{H}(\y_m, \y_M|\Dcal) \big), \label{eq:ac}
\end{align}
where $\lambda_m>0$ is the cost of querying a training example with fidelity $m$. When $m=M$, we have $a(\x, M) = \frac{1}{\lambda_M} \mathbb{H}\left(\y_M(\x)|\Dcal\right)$ ---  the output entropy.  Therefore, our acquisition function is an extension of the popular predictive entropy principle in conventional active learning.  At each step, we maximize our acquisition function to identify a pair of fidelity and input location that give the biggest benefit-cost ratio.  


\subsection{Efficient Acquisition Function Calculation}\label{sect:ac-compute}
To maximize the acquisition function \eqref{eq:ac}, a critical challenge is to compute the posterior of the outputs $\{\y_m\}$ in every fidelity, based on our model estimation results, \ie $p(\Wcal|\Dcal) \approx q(\Wcal)$.  Due to the high dimensionality of each $\y_m$ and the nonlinear coupling of the latent outputs across the fidelities (see \eqref{eq:our-model}), the computation is challenging and analytically intractable.  To address this issue, we consider approximating the posterior of the low dimensional latent output $\h_m$ first, which can be viewed as a function of the random NN weights $\bOmega_m = \{\W_1, \ldots, \W_m\}$ (given the input $\x$). We then use multivariate delta method~\citep{oehlert1992note,bickel2015mathematical} to compute the moments of $\h_m$. Specifically, we approximate $\h_m$ with a first-order Taylor expansion, 
\begin{align}
	\h_m(\bOmega_m) \approx \h_m(\EE[\bOmega_m]) + \J_m\left(\boldeta_m  - \EE[\boldeta_m]\right), \label{eq:1order}
\end{align}
where the expectation is under $q(\cdot)$,  $\boldeta_m = \vec(\bOmega_m)$, $\J_m = \frac{\partial \h_m}{\partial \boldeta_m}|_{\boldeta_m = \EE[\boldeta_m]}$ is the Jacobian matrix at the mean.  The rationale is as follows. First, $\h_m$ is linear to $\W_m$ and the second-order derivative is simply $\0$.  Second, as the NN output, $\h_m$ is highly nonlinear to the random weights in the previous  layers,  $\W_j$ ($j<m$). Hence, we can assume the change rate of $\h_m$ (\ie gradient or Jacobian) has much greater scales than the posterior covariance of $\W_j$ in the second-order term of the Taylor expansion. Note that $q(\W_j)$ is more informative and hence much more concentrated than the prior $ \N(\vec(\W_j)|\0, \I)$. The scale of the posterior covariance should be much smaller than $1$. Considering both cases, the first-order term can dominate the Taylor expansion and hence we ignore the higher order terms. Our ablation study has confirmed our analysis; see Sec. 4 of Appendix for details.   
Based on \eqref{eq:1order}, we can easily calculate the first and second moments of $\h_m$, 
\begin{align}
	\balpha_m &= \EE[\h_m] \approx \h_m\left(\EE[\bOmega_m]\right), 	\\
	 \V_m &= \text{cov}[\h_m] \approx \J_m \text{cov}(\boldeta_m) \J_m^\top, 
\end{align}
where  $\text{cov}(\boldeta_m) = \diag\left(\left\{\text{cov}\big(\vec(\W_j)\big)\right\}_{1\le j\le m}\right)$. 
Then we use moment matching to estimate a joint Gaussian posterior, $q(\h_m)  = \N(\h_m|\balpha_m, \V_m)$. Next, according to \eqref{eq:our-model}, we can obtain the posterior of the output $\y_m$, 
\begin{align}
	q(\y_m) = \N(\y_m|\A_m \balpha_m, \A_m \V_m \A_m^\top + \sigma_m^2\I).
\end{align}
The output entropy is  $\mathbb{H}(\y_m|\Dcal) = \frac{1}{2}\log|\A_m \V_m \A_m^\top + \sigma_m^2\I| + d_m \log\sqrt{2\pi e}$.  However, directly computing the log determinant of a $d_m \times d_m$ matrix is very expensive. To bypass this challenge, we use the Weinstein-Aronszajn identity~\citep{kato2013perturbation} to derive 
\begin{align}
	&\mathbb{H}(\y_m|\Dcal) =  \frac{1}{2}\log|\sigma_m^{-2}\A_m \V_m \A_m^\top + \I|  +  \log \left({2\pi e\sigma_m^2}\right)^{\frac{d}{2}} \notag \\
	&=\frac{1}{2}\log|\sigma_m^{-2}\A_m^\top\A_m\V_m + \I| + d_m \log \sqrt{2\pi e\sigma_m^2}. \label{eq:en_ym}
\end{align}
Now, the log determinant is calculated on a much smaller, $k_m \times k_m$ matrix, which is very cheap and efficient. 

Using a similar approach, we can calculate the joint posterior of $\barf_m = [\h_m; \h_M]$ and then $\bary_m=[\y_m; \y_M]$.  First, we view $\barf_m$ as a function of all the random weights, $\Wcal = \{\W_1, \ldots, \W_M\}$. We use the multivariate delta method and moment matching to estimate a joint Gaussian posterior, 
$q(\barf_m) = \N(\barf_m|\baralpha_m, \barV_m)$ where $\baralpha_m = \barf_m\left(\EE[\Wcal]\right)$,  $\barV_m = \barJ_m \text{cov}(\boldeta)\barJ_m^\top$, $\boldeta = \vec(\Wcal)$, $\barJ_m = \frac{\partial \barf_m}{\partial \boldeta}|_{\boldeta = \EE[\boldeta]}$, and $\text{cov}(\boldeta) = \diag\left(\left\{\text{cov}\big(\vec(\W_j)\big)\right\}_{1\le j\le M}\right)$. According to our model \eqref{eq:our-model}, we can represent
\[
\bary_m = \barA_m \barf_m + \bareps_m,
\]
where $\barA_m = \diag(\A_m, \A_M)$ and $\bareps_m = [\bepsilon_m; \bepsilon_M]$. Therefore, the joint posterior of $\bary_m$ is 
\begin{align}
	q(\bary_m) = \N(\bary_m|\barA_m \baralpha_m, \barA_m \barV_m \barA_m^\top + \S_m),
\end{align}
where $\S_m = \diag(\sigma^2_m\I_{d_m}, \sigma^2_M \I_{d_M})$,  $\I_{d_m}$ and $\I_{d_M}$ are identity matrices of $d_m \times d_m$ and $d_M \times d_M$, respectively.  Again, we use  Weinstein-Aronszajn identity to simplify the entropy computation of $\bary_m$ (\ie $\y_m$ and $\y_M$), 
\begin{align}
	&\mathbb{H}(\y_m, \y_M|\Dcal)  =\frac{1}{2}\log|\S_m^{-1}\barA_m \barV_m \barA_m^\top + \I|  + \beta_m \notag \\
	&=\frac{1}{2}\log|\barA_m^\top \S_m^{-1} \barA_m \barV_m + \I| + \beta_m, \label{eq:en_ymM}
\end{align}
where $\beta_m = d_m \log \sqrt{2\pi e\sigma_m^2} + d_M \log \sqrt{2\pi e\sigma_M^2}$ and the log determinant is computed from a $(k_m + k_M) \times (k_m + k_M)$ matrix, which is cheap and efficient. Note that we can use block matrices to compute $\barA_m^\top \S_m^{-1} \barA_m = \diag\left(\sigma_m^{-2}\A_m^\top \A_m, \sigma_M^{-2}\A_M^\top \A_M\right)$.

Now, based on \eqref{eq:en_ym} and \eqref{eq:en_ymM}, we can calculate our acquisition function \eqref{eq:ac} in an analytic and deterministic way. For each fidelity $m$, we maximize $a(\x, m)$ w.r.t to $\x$ to find the optimal input. We can use automatic differential libraries to calculate the gradient and feed it to any optimization algorithm, \eg L-BFGS.   We then use the optimal input at the fidelity that has the largest acquisition function value to query  the next training example. Our deep multi-fidelity active learning is summarized in Algorithm \ref{alg:dmf-al}. 
\setlength{\textfloatsep}{20pt}
\begin{algorithm}                      
	\small
	\caption{\ours($\Dcal$, $T$, $\{\lambda_m\}_{m=1}^M$ )}          
	\label{alg:dmf-al}                           
	\begin{algorithmic}[1]                    
		\STATE Train the deep multi-fidelity model \eqref{eq:joint-prob} on the initial dataset $\Dcal$ with stochastic structural variational learning.  
		\FOR {$t=1,\ldots, T$}
		\STATE Based on \eqref{eq:en_ym} and \eqref{eq:en_ymM}, calculate and optimize the acquisition function \eqref{eq:ac} to find 
		\begin{align}
			(\x_{t}, m_{t}) = \argmax_{\x \in \Xcal, 1\le m \le M} a(\x, m).\notag 
		\end{align}
		\STATE Query the output $\y_{t}$ at input $\x_t$ with fidelity $m_t$. 
		\STATE $\Dcal \leftarrow \Dcal \cup \{(\x_{t}, \y_{t})|m_t\}$.
		\STATE Re-train the deep multi-fidelity model on $\Dcal$.
		\ENDFOR
	\end{algorithmic}
\end{algorithm}
 \vspace{-0.05in}
\subsection{Algorithm Complexity}
\vspace{-0.1in}
The time complexity of training our deep multi-fidelity model is $\Ocal(N(\sum_{m=1}^M (k_ml_m)^2 + F))$ where $N$ and $F$ are the total number of training examples and NN parameters, respectively. The space complexity is $\Ocal(F + \sum_{m=1}^M (k_ml_m)^2)$, which is to store the NN parameters, and posterior mean and covariance of each random weight matrix $\W_m$. The time complexity of calculating the acquisition function is $\Ocal(\sum_{m=1}^M d_mk_m^2 + k_m^3)$. Since $k_m \ll d_m$, the complexity is linear in the output dimensions. Due to the usage of the learned variational posterior $q(\Wcal)$, the space complexity of computing the acquisition function is the same as that of training the multi-fidelity model.

\vspace{-0.15in}
\section{Related Work}
\vspace{-0.2in}
Active learning (AL) is a fundamental machine learning topic~\citep{balcan2007margin,settles2009active,balcan2009agnostic,dasgupta2011two,hanneke2014theory}.
Recent research focuses more on deep neural networks.  \citet{gal2017deep} used Monte-Carlo (MC) Dropout~\citep{gal2016dropout} to perform variational inference for Bayesian neural networks. The dropout samples can be viewed as the posterior samples of the output. These samples are then used to compute an information measure, such as predictive entropy and BALD~\citep{houlsby2011bayesian}, to select unlabeled examples to query. This method has achieved a great success in image classification. Following this work, \citet{kirsch2019batchbald} developed a greedy approach to select a batch of unlabeled examples each time, so as to improve the efficiency of active learning. 
There are also other excellent works along this line. For example, \citep{geifman2017deep,sener2018active} selected representative examples based on core-set search. \citep{gissin2019discriminative} selected maximally indistinguishable samples from the unlabeled pool, and the idea is reminiscent of generative adversarial networks~\citep{goodfellow2014generative}.  \citep{ducoffe2018adversarial} used adversarial samples to calculate the distance to the decision boundary and select examples accordingly to label. \citep{ash2019deep} measured the uncertainty in terms of the gradient magnitude and select a disparate batch of inputs in a hallucinated gradient space. 

Our work differs from the existing studies in several aspects. First, most methods are pool-based active learning, \ie a pool of unlabeled examples are collected beforehand, and new examples are only selected from the pool. This is reasonable when the training input is high-dimensional or hard to generate, \eg image classification. By contrast, our work focuses on learning mappings from low-dimensional inputs to high-dimensional outputs, which are common in physical simulation and engineering design. To find the best training example, we \textit{optimize} the acquisition function in the entire domain rather than limit the search in a discrete set.  Second, most active learning methods assume a uniform fidelity (or quality) of the training examples. Our work considers the case that the training examples can be queried with multiple fidelities, resulting in different cost/quality trade-offs. Our goal is to maximize the learning improvement while minimizing the query cost, \ie maximizing the benefit-cost ratio.  This naturally extends the standard active learning, where the total cost is proportional to the number of queries due to the uniform fidelity. The pioneer work of \citep{settles2008active}
empirically studies active learning with nonuniform human  labeling costs.  To our knowledge, our work is the first multi-fidelity active learning approach for  high-dimensional outputs. 

Many multi-fidelity (MF) models have been proposed, while their active training approaches are lacking. These models are often based on Gaussian processes (GPs). For example, \citet{perdikaris2017nonlinear} successively estimated a set of GPs, where each GP predicts the output for one fidelity, where the input comes from the original input and the output of the previous fidelity. In \citep{cutajar2019deep}, this cascade model  is jointly estimated. \citet{hamelijnck2019multi} used Gaussian process regression networks (GPRN) ~\citep{wilson2012gaussian}, deep GP~\citep{damianou2013deep} and mixture of experts~\citep{rasmussen2002infinite} to develop a multi-task, multi-resolution GP model. Despite their success, these models do not scale to high-dimensional outputs. The most recent work by ~\citet{wang2021multi} overcomes this limitation by proposing a nonlinear coregionalization component and then stacking these components for multi-fidelity modeling. How to conduct active learning for this model, however,  remains an open problem --- even considering our proposed acquisition function, the computation is challenging due to the complicated couplings of the kernels and weight functions. In our experiments, we have compared the performance of our model with \citep{wang2021multi}, and other state-of-the-art high-dimensional GP regressors (that are single-fidelity) in non-active learning. The results confirmed  the advantage of our NN based multi-fidelity model (see Sec.  3 of Appendix). Note that GPs are often used for learning physics and experiment design~\citep{sacks1989design}. 

AL also connects to Bayesian optimization (BO)~\citep{mockus2012bayesian,snoek2012practical}, while the latter aims to find the function optimum and so the acquisition function definition, computation, and optimizing techniques/challenges are very different. Interestingly, while multi-fidelity AL methods are scarce, MF BO is  common, \eg   
\citep{huang2006sequential, lam2015multifidelity,picheny2013quantile,kandasamy2016gaussian,kandasamy2017multi,poloczek2017multi,mcleod2017practical,wu2018continuous,song2019general,takeno2019multi}.  Most MFBO approaches are based on GPs --- the classical BO framework.  The recent MFBO work of \citet{li2020multi} uses a chain of NNs to estimate a multi-fidelity surrogate of the black-box objective function, queries new examples based on the maximum-value entropy search (MES) principle~\citep{wang2017max}.  From the modeling perspective, its key difference (from our model) is that the NN output at each fidelity is directly fed into the NN for the next fidelity. Therefore, their model needs massive weight parameters to connect multiple NNs for high-dimensional outputs, which can be very inefficient. In addition, the massive outputs of the previous fidelity might dominate the original input (low-dimensional) to the model. Furthermore, \citet{li2020multi} developed a recursive one-dimensional  quadrature to estimate the posterior of the output at each fidelity. While being suitable for optimizing a single-output black-box function, this technique is difficult to extend to multiple outputs (due to the explosion of the computational cost for multi-dimensional quadrature), not to mention a large number of  outputs in our active learning setting. 
\section{Experiment}
\vspace{-0.1in}
\subsection{Solving Partial Differential Equations}
\vspace{-0.1in}
We first evaluated \ours in standard computational physics tasks. Specifically, we used \ours to predict the solution fields of  three commonly used partial differential equations (PDEs): \textit{Burgers}' , \textit{Poisson}'s and \textit{Heat} equations~\citep{olsen2011numerical}. The training examples are collected by running a numerical solver with different meshes. The more the nodes/steps to create the mesh, the higher the fidelity. The input includes the PDE parameters and/or parameterized initial or boundary conditions. The output consists of the solution values at the mesh used in the solver. For example, a $50 \times 50$ mesh corresponds to a $2,500$ dimensional output vector.  For active learning, we considered two-fidelity queries for all the three equations, where the sizes of the corresponding output fields are $16 \times 16$ and $32 \times 32$. In addition, we considered a three-fidelity setting for Poisson's equation, denoted by \textit{Poisson-3}, and the sizes of the output fields of the three fidelities are $16 \times 16$ , $32 \times 32$, and $64 \times 64$, respectively. For the two-fidelity active learning, we uniformly sampled the inputs, and queried $10$ training examples at the first fidelity and $2$ at the second fidelity. We used those examples as the initial training dataset. Similarly, for \textit{Poisson-3}, we collected $10$, $5$ and $2$ examples in the first, second, and third fidelity as the initial training set. We generated $500$ samples for test, where the test inputs were uniformly sampled from the domain. The outputs are calculated by running the solver with an even denser mesh --- $128 \times 128$ for Burger's and Poisson's equations, and $100 \times 100$ for Heat equation --- and interpolating the solution values at the target grid~\citep{zienkiewicz1977finite} (this is the standard approach in physical simulation and  the accuracy does not change). More details are given in the Appendix. \textit{We ran the solvers at each fidelity for many times and calculated the average running time. We then normalized the average running time to obtain $\lambda_1 = 1$, $\lambda_2 = 3$ and $\lambda_3 = 10$}. 

\noindent\textbf{Competing methods.}  We compared \ours with the following active learning approaches. (1) {MF-BALD},   a straightforward extension of BALD~\citep{houlsby2011bayesian} to integrate multi-fidelity queries. The acquisition function is $a_{\text{MF-BALD}}(\x, m) = \frac{1}{\lambda_m} \mathbb{I}\big(\y_m(\x), \Wcal|\Dcal\big) = \frac{1}{\lambda_m}\left(\mathbb{H}(\y_m(\x)|\Dcal) - \EE_{p(\Wcal|\Dcal)}\left[\mathbb{H}(\y_m(\x)|\Wcal, \Dcal)\right]\right) = \frac{1}{\lambda_m}\left(\mathbb{H}(\y_m(\x)|\Dcal) - \frac{d_m}{2}\log(2\pi e \sigma_m^2)\right)$. 
Note that conditioned the NN parameters, the entropy of the observed output $\y_m$ is only determined by the noise variance $\sigma_m^2$.
(2) {MF-PredVar}, a straightforward extension of the popular predictive variance principle, $a_{\text{MF-PredVar}} = \frac{1}{\lambda_m} \frac{1}{d_m}\sum_{j=1}^{d_m} \text{Var}(y_{mj}|\Dcal)$. (3) {Dropout-latent}, where we use MC dropout~\citep{gal2017deep} for variational inference, each time draw $100$ dropout samples for the low dimensional latent outputs $\{\h_1(\x), \ldots, \h_M(\x)\}$ to estimate multi-variate Gaussian posteriors for each $\h_m$ and $\barf_m = [\h_m; \h_M]$ (via empirical means and covariance matrices), and then follow Section \ref{sect:ac-compute} to calculate and optimize the acquisition function \eqref{eq:ac}. Note that we have also used MC dropout to outright sample the final, high-dimensional outputs $\{\y_m\}$ and estimate multi-variate Gaussian posteriors to calculate \eqref{eq:ac} and $a_{\text{MF-BALD}}$. While doing this is much more expensive, the performance did not improve. Instead, it was way worse than {Dropout-latent} and {MF-BALD}. See the details in the Appendix. (4) {MF-Random}, where each time we randomly select a fidelity and then an input  to query. 
(5) {Random-F1}, (6) {Random-F2}, and (7) {Random-F3}, where we stick to the first, second and third fidelity, respectively, and randomly sample an input to query each time.

\begin{figure*}
	\centering
	\setlength\tabcolsep{0pt}
	\begin{tabular}[c]{ccc}
		\setcounter{subfigure}{0}
		\begin{subfigure}[t]{0.33\textwidth}
			\centering
			\includegraphics[width=\textwidth]{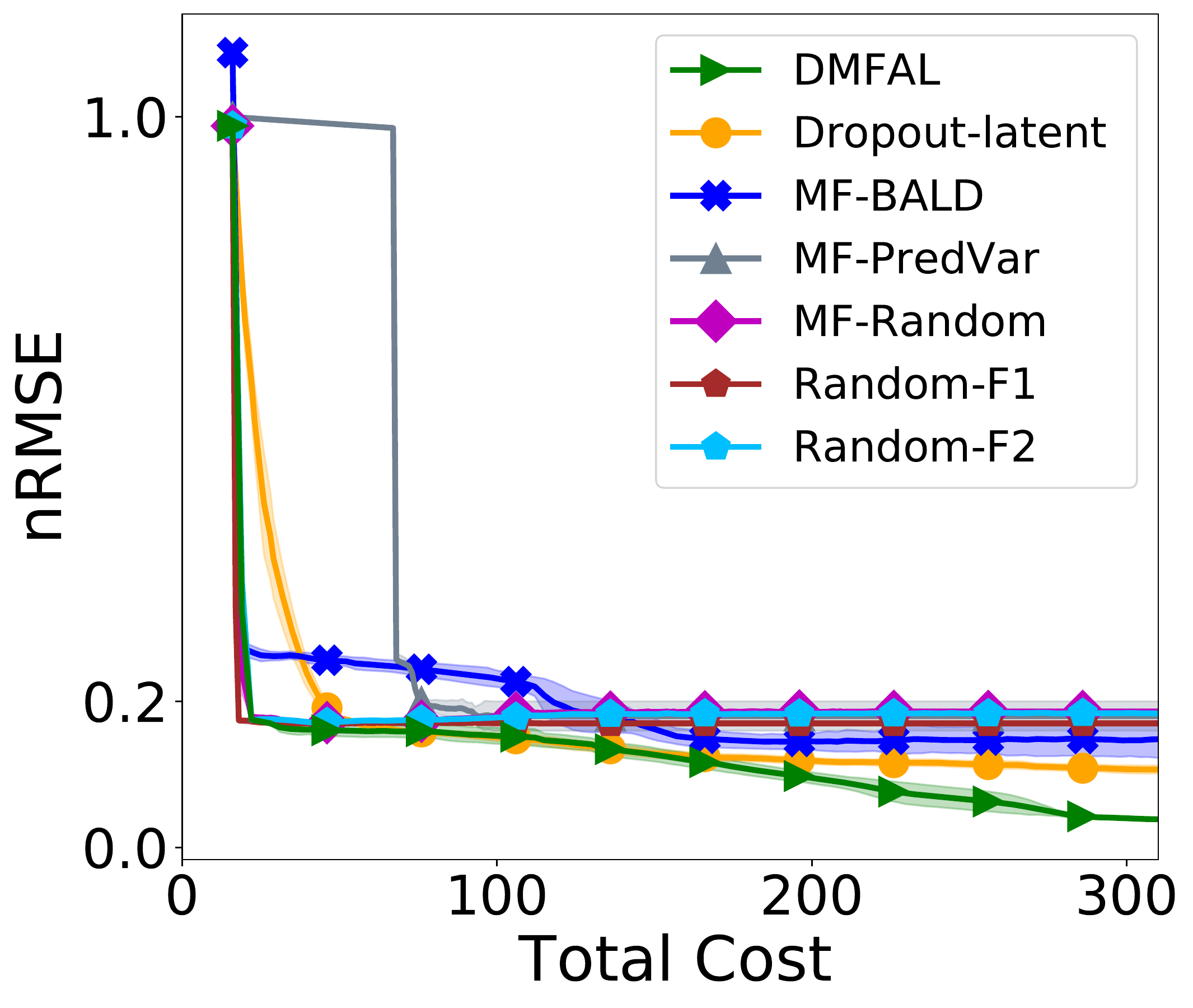}
			\caption{\small \textit{Burgers}}
		\end{subfigure} &
		\begin{subfigure}[t]{0.33\textwidth}
			\centering
			\includegraphics[width=\textwidth]{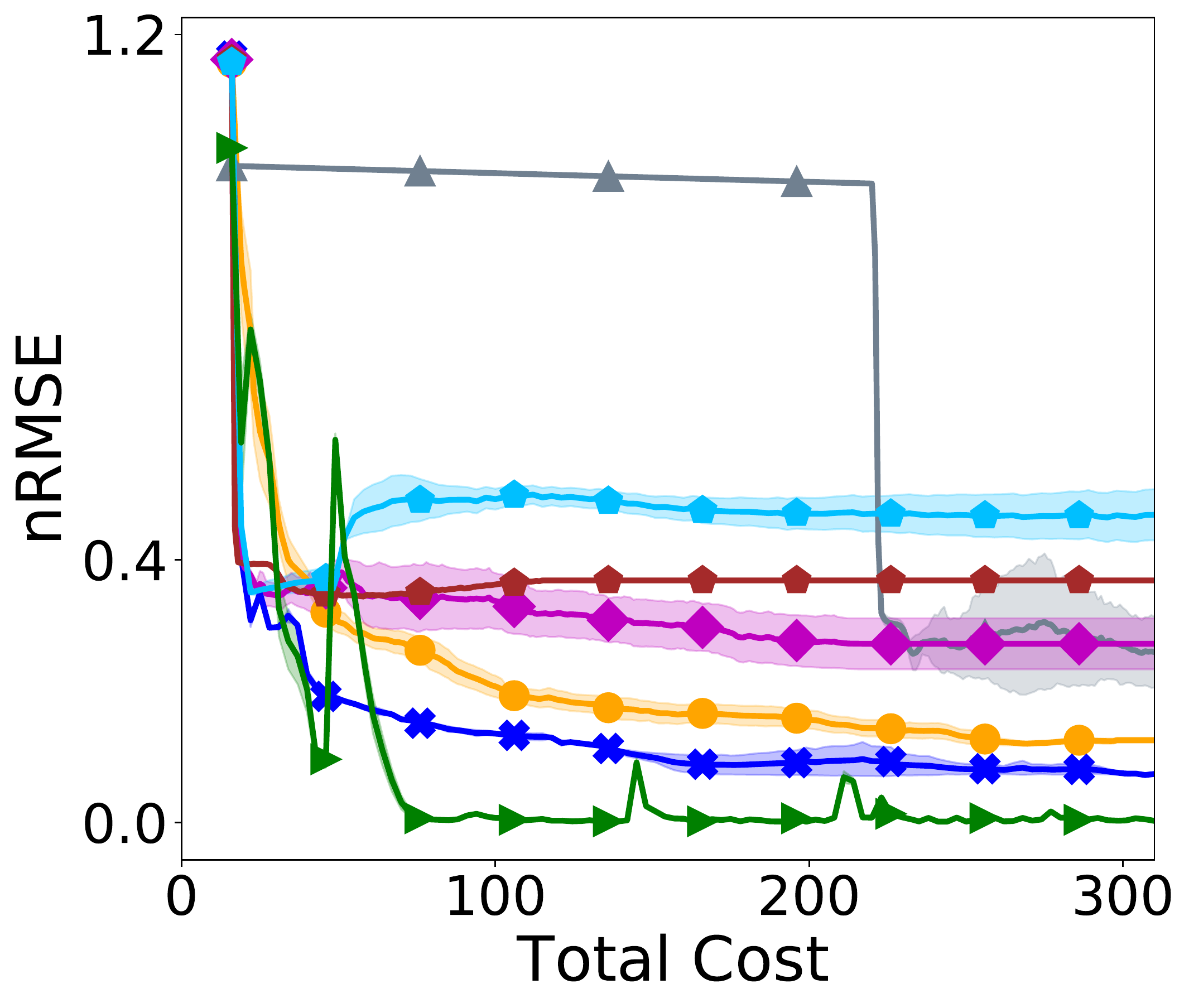}
			\caption{\small \textit{Heat}}
		\end{subfigure}
		&
		\begin{subfigure}[t]{0.33\textwidth}
			\centering
			\includegraphics[width=\textwidth]{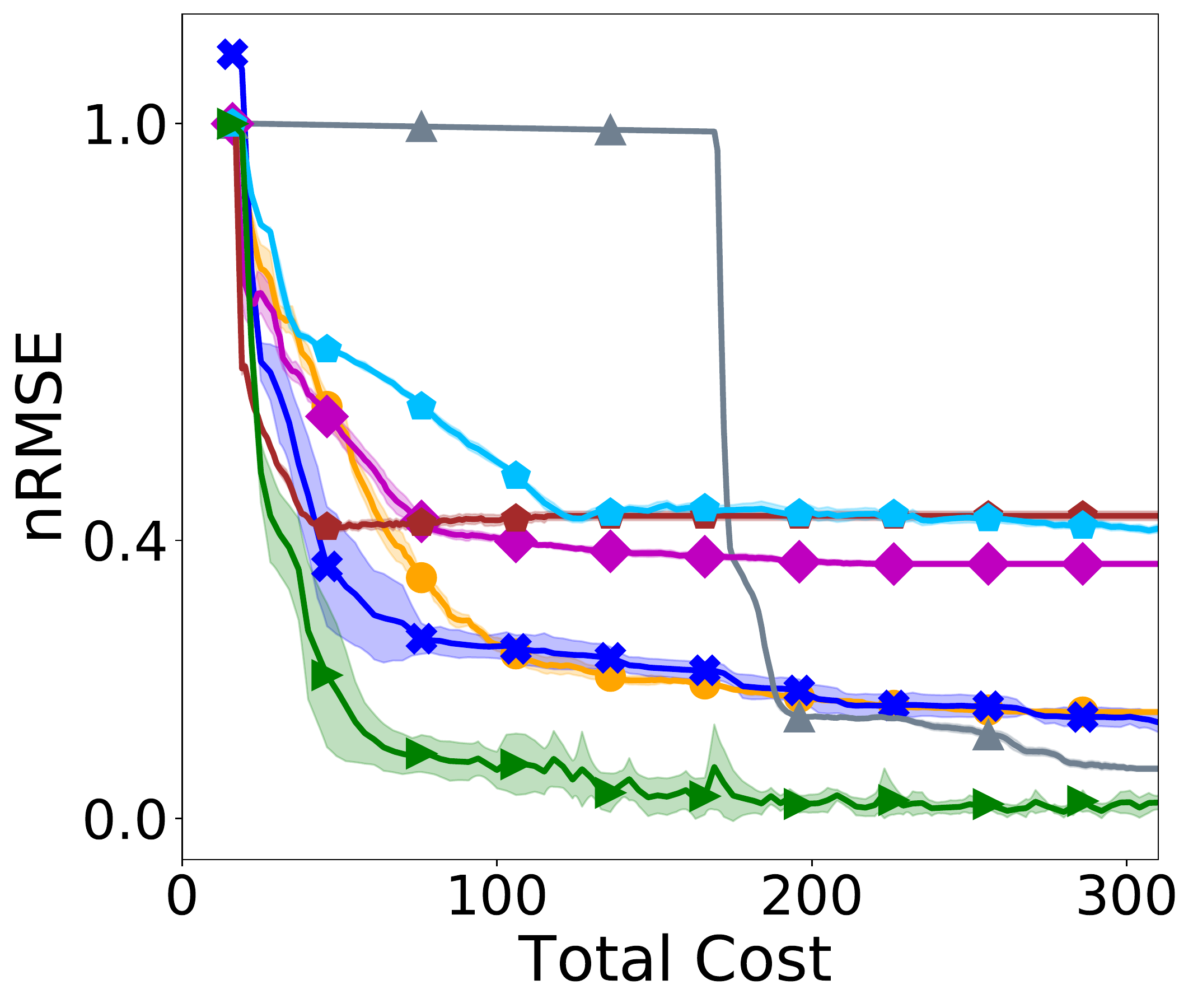}
			\caption{\small \textit{Poisson}}
		\end{subfigure}\\
		\begin{subfigure}[t]{0.33\textwidth}
			\centering
			\includegraphics[width=\textwidth]{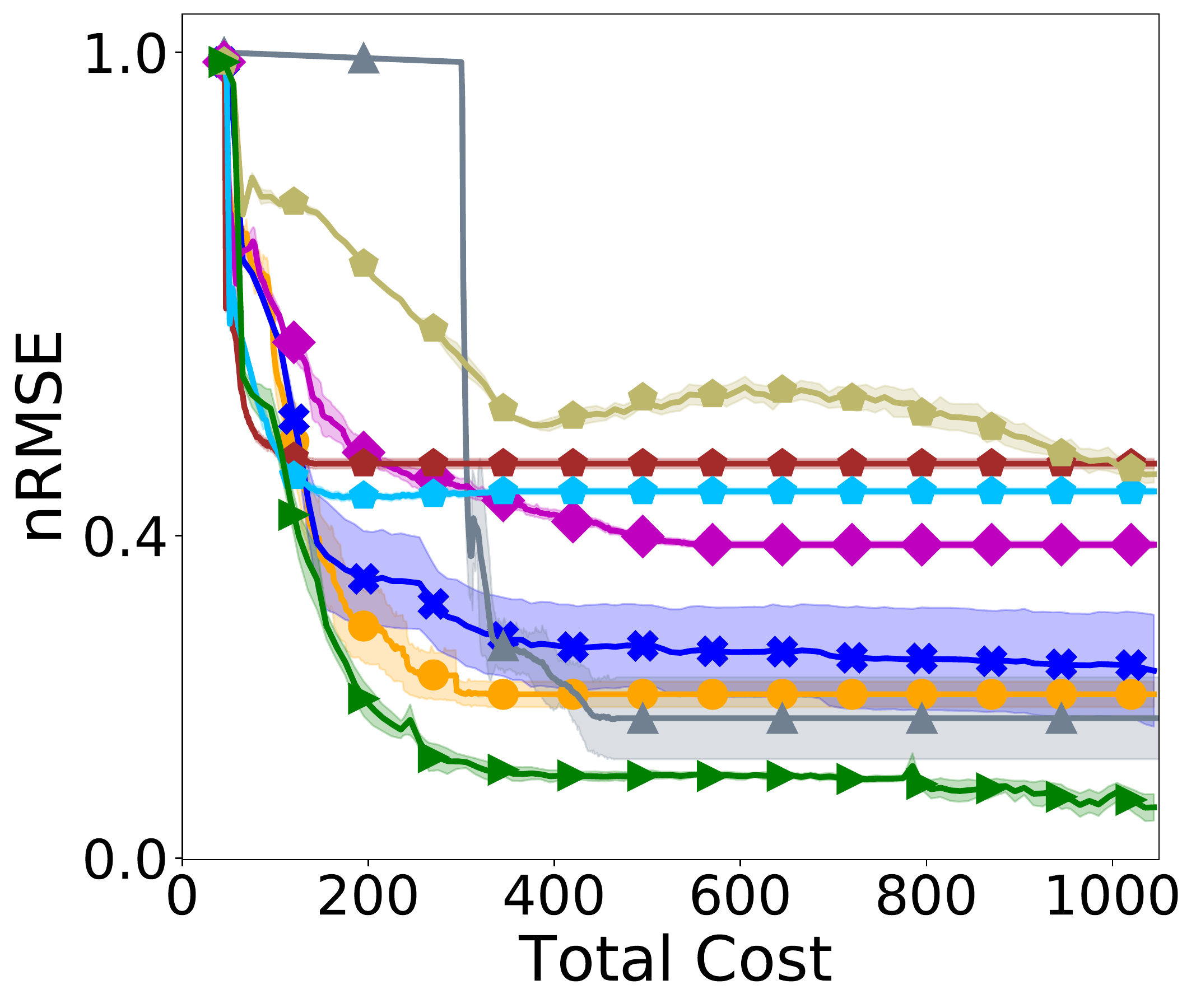}
			\caption{\small \textit{Poisson-3}}\label{fig:solving-po-3fid}
		\end{subfigure} &
		\begin{subfigure}[t]{0.33\textwidth}
			\centering
			\includegraphics[width=\textwidth]{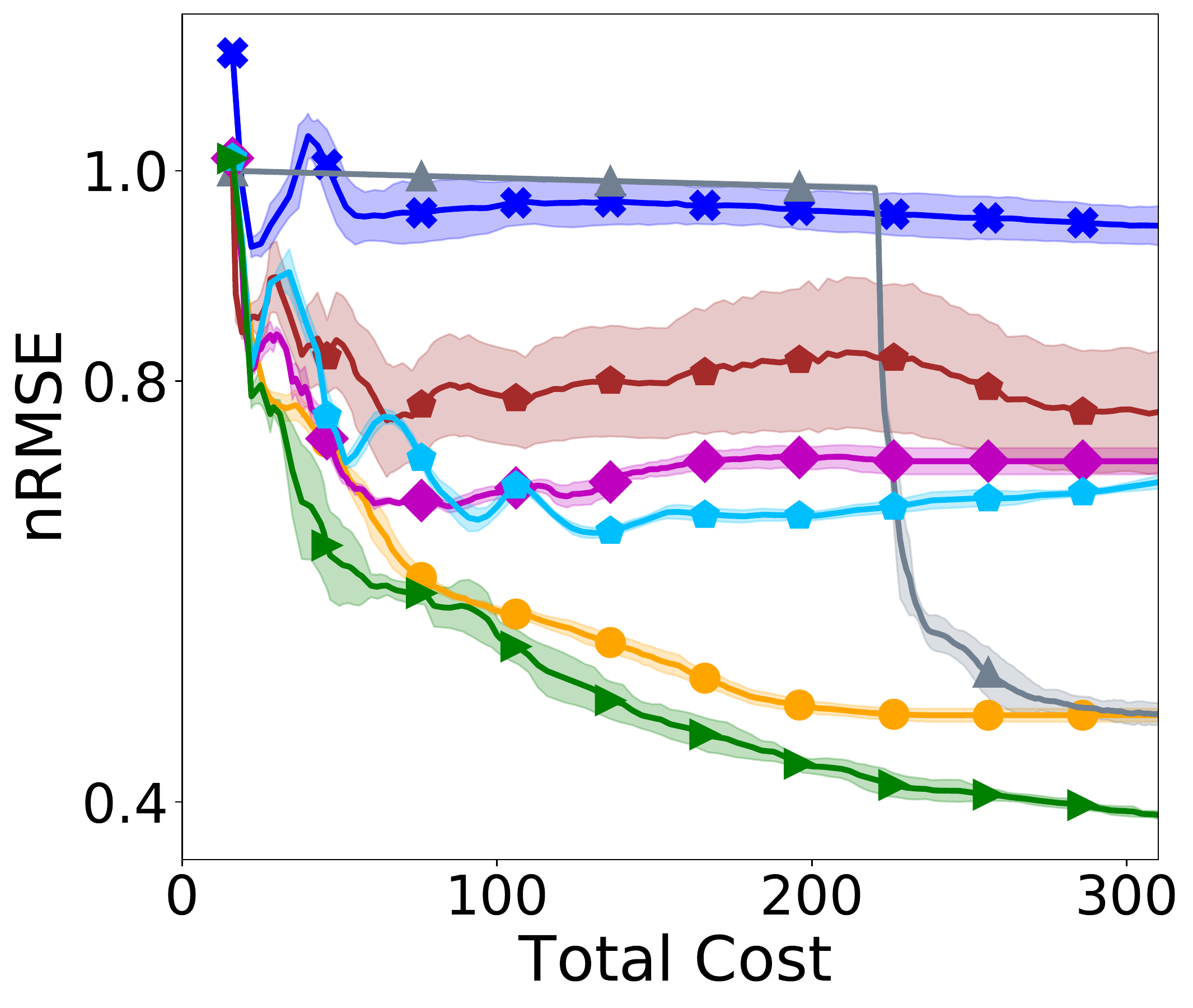}
			\caption{\small \textit{Topology structure optimization}} \label{fig:tpo}
		\end{subfigure}
		&
		\begin{subfigure}[t]{0.33\textwidth}
			\centering
			\includegraphics[width=\textwidth]{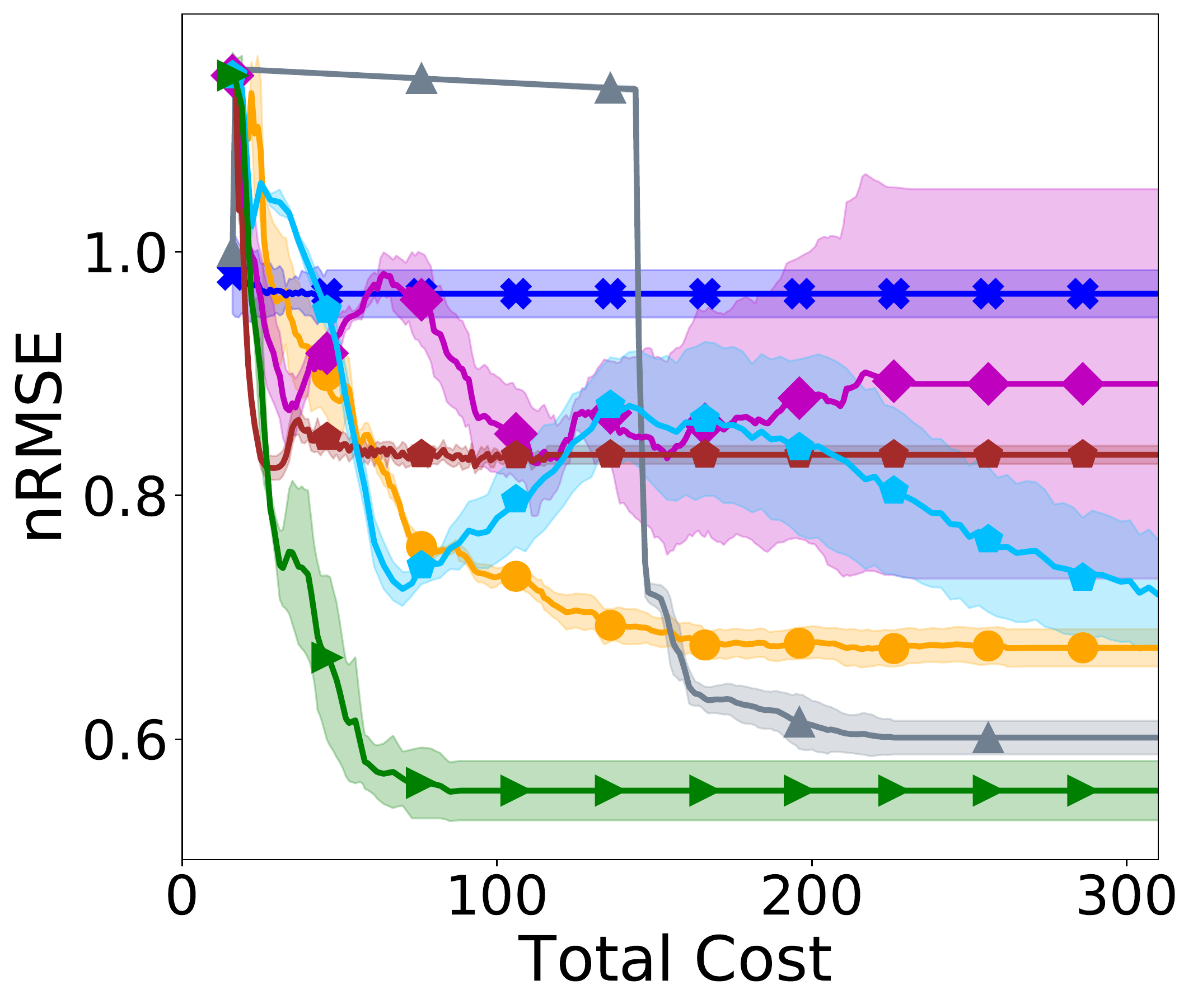}
			\caption{\small \textit{Fluid dynamics}}\label{fig:ns}
		\end{subfigure}
	\end{tabular}
	\vspace{-0.1in}
	\caption{\small Normalized root-mean-square error (nRMSE) \textit{vs.} total cost (running time) in active learning. The results are averaged from five runs. The shaded regions indicate the standard deviations.} \label{fig:al-acc}
	\vspace{-0.2in}
\end{figure*}

\noindent\textbf{Settings and results.} We introduced a two-layer NN for each fidelity and used \texttt{tanh} as the activation function\footnote{We tried more layers per fidelity, \ie 3-5, and the overall performance did not improve. In addition, our experience shows that alternative activation functions, like \texttt{ReLU} and \texttt{LeakyReLu}, are inferior to \texttt{tanh}. This is consistent with the choice of typical physics informed neural networks~\citep{raissi2019physics}.}. The layer width  was chosen from $\{8, 16, 32, 64, 128\}$.  We set the same dimension for the latent output in each fidelity and selected it from $\{5, 10, 15, 20\}$. For {Dropout-latent}, we tuned the dropout rate from $\{0.1, 0.2, 0.3, 0.4, 0.5\}$. All the methods were implemented with PyTorch~\citep{paszke2019pytorch}. We used ADAM~\citep{kingma2014adam} for stochastic optimization, where the learning rate was selected from $\{10^{-4}, 5 \times 10^{-4}, 10^{-3}, 5 \times 10^{-3}, 10^{-2}\}$.  We set the number of epochs to $2,000$, which is enough for convergence. We conducted five runs for each method, and in each run, we queried $100$ examples. We report the average normalized root-mean-square-error (nRMSE) \textit{vs.} the accumulated cost in Fig.  \ref{fig:al-acc}a-d. The shaded region shows the standard deviation. We can see that at the beginning, all the methods have the same or comparable performance. Along with more queries, \ours quickly achieves better prediction accuracy, and continues to outperform all the other methods  by a large margin. Therefore, \ours can perform much better with the same cost or achieve the same performance with the smallest cost. It is interesting to see that in Fig. \ref{fig:al-acc}a, while all the competing approaches have saturated early, \ours keeps improving its prediction accuracy. 
 \setlength{\columnsep}{5pt}
\begin{figure}[!htb]
	\centering
	\begin{subfigure}[b]{0.4\textwidth}
		\centering
		\includegraphics[width=1.0\linewidth]{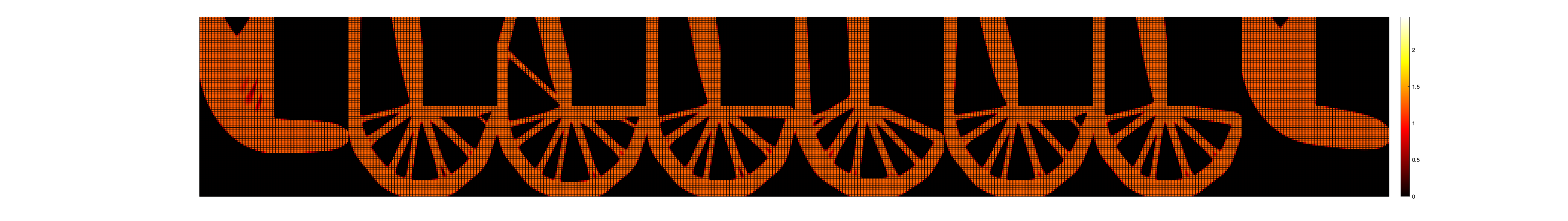}
		\caption{\small Ground Truth}
		\label{fig:tpo-gt}
	\end{subfigure}
	\begin{subfigure}[b]{0.4\textwidth}
		\centering
		\includegraphics[width=1.0\linewidth]{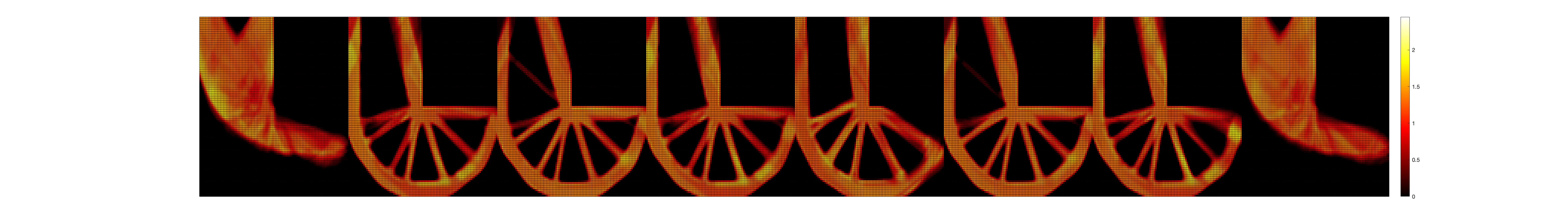}
		\caption{\small \ours}
		\label{fig:tpo-ours}
	\end{subfigure}
	\begin{subfigure}[b]{0.4\textwidth}
		\centering
		\includegraphics[width=1.0\linewidth]{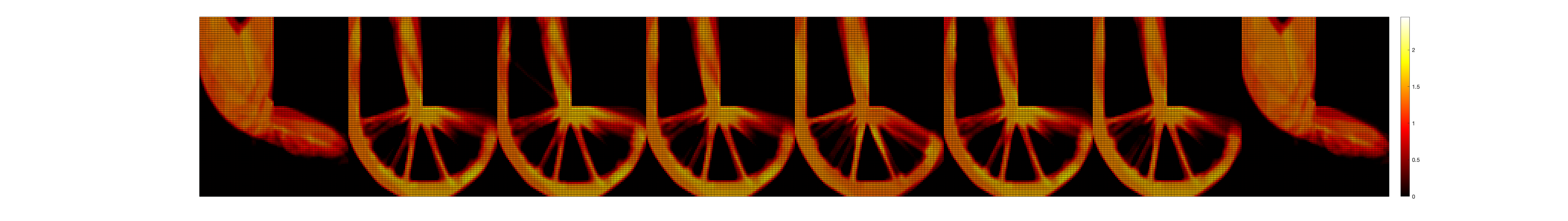}
		\caption{\small Dropout-latent}
		\label{fig:tpo-dropout}
	\end{subfigure}
	\begin{subfigure}[b]{0.4\textwidth}
		\centering
		\includegraphics[width=1.0\linewidth]{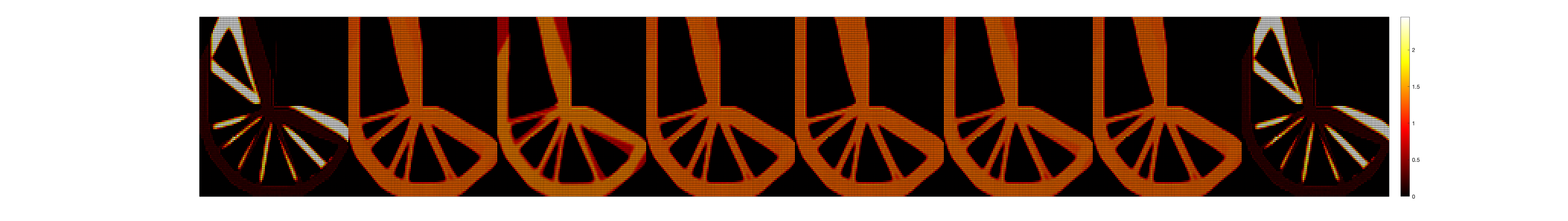}
		\caption{\small MF-BALD}
		\label{fig:tpo-bald}
	\end{subfigure}
	\begin{subfigure}[b]{0.4\textwidth}
	\centering
	\includegraphics[width=1.0\linewidth]{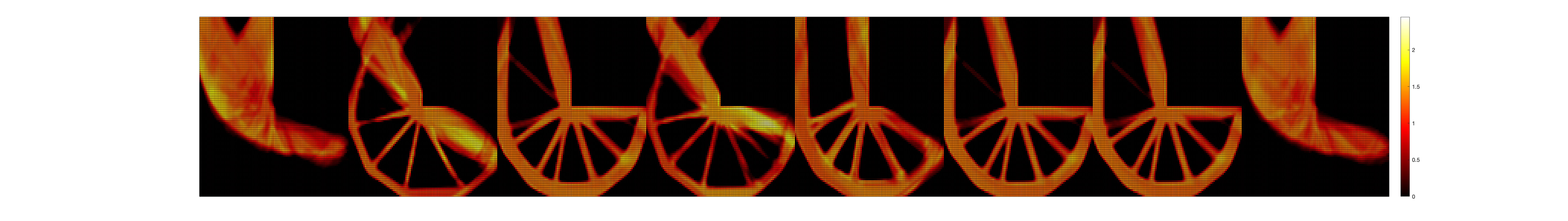}
	\caption{\small MF-PredVar}
	\label{fig:tpo-pdv}
	\end{subfigure}
	\begin{subfigure}[b]{0.4\textwidth}
		\centering
		\includegraphics[width=1.0\linewidth]{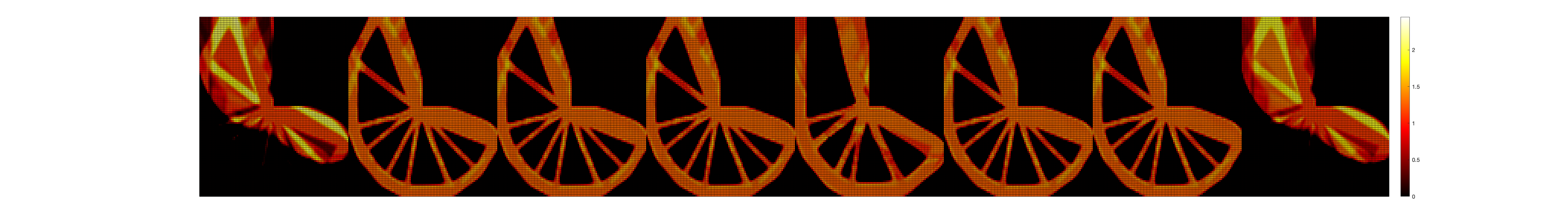}
		\caption{\small MF-Random}
		\label{fig:tpo-random}
	\end{subfigure}
	\begin{subfigure}[b]{0.4\textwidth}
		\centering
		\includegraphics[width=1.0\linewidth]{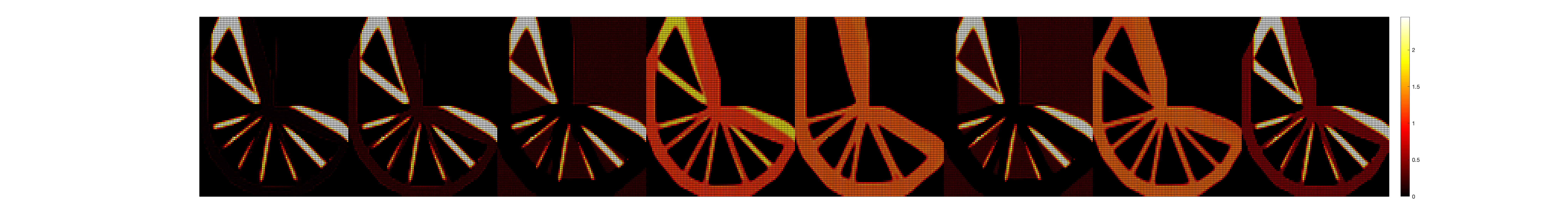}
		\caption{\small Random-F1}
		\label{fig:tpo-random-1}
	\end{subfigure}
	\begin{subfigure}[b]{0.4\textwidth}
		\centering
		\includegraphics[width=1.0\linewidth]{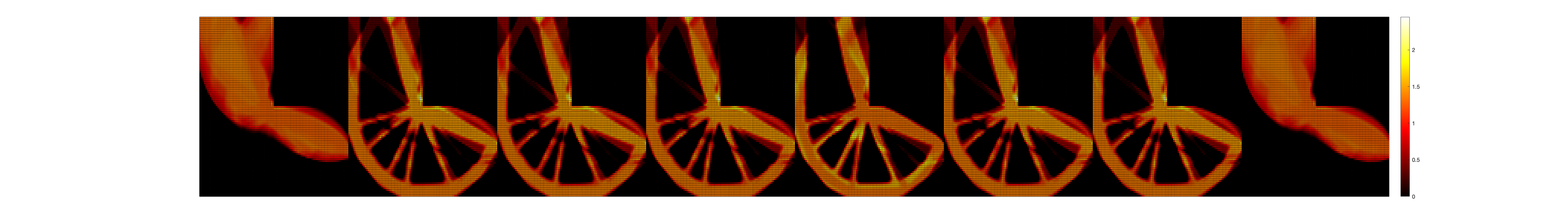}
		\caption{\small Random-F2}
		\label{fig:tpo-random-2}
	\end{subfigure}
	\vspace{-0.1in}
	\caption{\small The predicted topology structures for $8$ loads. All the active learning approaches started with the same training set (10 fidelity-1 and 2 fidelity-2 examples), and ran with $100$ queries.}
	\label{fig:pred-top-structure}
	\vspace{-0.2in}
\end{figure}

 The inferior performance of {MF-BALD} and {MF-PredVar} to \ours indicates that the straightforward extension of BALD and predictive variance can be suboptimal, which  might partly because they miss the influence of the low-fidelity training examples on the final predictions (\ie at the highest fidelity). How to better generalize BALD and PredVar to the multi-fidelity setting remains an open problem. 
  Overall,  these results have demonstrated the advantage of our deep multi-fidelity active learning  approach. 
\vspace{-0.1in}
\subsection{Topology Structure Optimization}
\vspace{-0.1in}
Next, we applied \ours in topology structure optimization.  A topology structure is a layout of materials, \eg alloy and concrete, in some designated spatial domain. Given the input from the outside environment, \eg external force, we want to find an optimal structure that achieves the maximum (or minimum)  interested property, \eg stiffness. Topology structure optimization is crucial to many engineering design and manufacturing problems, including 3D printing, and design of air foils, slab bridges, aerodynamic shapes of race cars, \etc The conventional approach is to solve a constraint optimization problem that minimizes a compliance objective subject to a total volume constraint~\citep{sigmund1997design}. However, the numerical computation is usually very costly. We aim to use active learning to learn a model that directly predicts the optimal structure, without the need for running numerical optimization every time. 

We considered the stress experiment in~\citep{keshavarzzadeh2018parametric} with an L-shape linear elastic structure.  The structure is subjected to a load (\ie input) on the bottom right half and is discretized in a $[0, 1] \times [0, 1]$ domain. The load is represented by two parameters, the location (in $[0.5, 1]$) and angle (in $[0, \frac{\pi}{2}]$). The goal is to find the structure that achieves the maximum stiffness given the load. Optimizing the structure  needs to repeatedly call a numerical solver, where the choice of the mesh determines the fidelity.  We used two fidelities to query the training examples. One uses a $50 \times 50$ mesh, the other $75 \times 75$. Correspondingly, the output dimensions are $2,500$ and $5,625$. The costs are measured by the average running time: $\lambda_1 = 1, \lambda_2 = 3$. We randomly generated $500$ structures for test, where the internal solver uses a $100 \times 100$ mesh. Initially, we randomly queried $10$ examples at the first fidelity and $2$ at the second fidelity. We then ran all the active learning methods to query $100$ examples. We conducted the experiments for five times, and report the average nRMSE  along with the cost in  Fig. \ref{fig:tpo}. As we can see, \ours achieves much better prediction accuracy than the competing approaches (with the same cost). That implies our predicted structures are much closer to the optimal structures.  While the performance of the other methods tended to converge early, \ours 's performance kept improving and the trend did not stop even when all the queries were finished. 
  
For a fine-grained comparison, we visualize eight structures predicted by all the methods, after the active learning is finished. As shown in Fig. \ref{fig:pred-top-structure}, \ours predicted much more accurate structures, which capture both the global shapes and local details, and the density of the materials is closer to the ground-truth.  Although {Dropout-latent} captures the global shapes as well, its predictions are more blurred and miss many local details, \eg the second to seventh structure in Fig. \ref{fig:pred-top-structure}c. The other methods often provided wrong structures (\eg the first and last structure in Fig. \ref{fig:pred-top-structure}d, f, and g, the second and fourth structure in Fig. \ref{fig:pred-top-structure}e) and insufficient density (\eg the second, third and sixth structure in \ref{fig:pred-top-structure}g).

\vspace{-0.15in}
\subsection{Predicting Fluid Dynamics}
\vspace{-0.1in}
Third, we applied \ours in a computational fluid dynamics (CFD) problem.  The task is to predict the first component of the velocity field of a flow within a rectangular domain in $[0, 1] \times [0, 1]$. The flow is driven by the boundaries with a prescribed velocity (\ie input)~\citep{bozeman1973numerical}.
Along with time, the local velocities inside the fluid will vary differently, and eventually result in turbulent flows. Computing these fields along with time requires us to solve the incompressive Navier-Stokes equations~\citep{chorin1968numerical}, which is known to be challenging to solve because of their complex behaviors under big Reynolds numbers. We considered active learning with two-fidelity queries to predict the first component of the velocity field  at evenly spaced $20$ time points in $[0, 10]$ (temporal domain). The examples in the first fidelity were generated with a $50 \times 50$ mesh in the spatial domain (\ie $[0, 1]\times [0, 1]$), and the second fidelity $75 \times 75$. The corresponding output dimensions are $50,000$ and $112,500$. The input is a five dimensional vector that consists of the prescribed boundary velocity and Reynold number. See more details in the Appendix. To collect the test dataset, we randomly sampled $256$ inputs and computed the solution with a $128 \times 128$ mesh. The test outputs are obtained by the cubic-spline interpolation. At the beginning, we randomly queried $10$ and $2$ training examples in the first and second fidelity, respectively. Then we ran each active learning method with $100$ queries. We repeated the experiments for five times. The average nRMSE along with the accumulated cost is shown in Fig. \ref{fig:ns}. It can be seen that during the training, \ours consistently outperforms all the competing methods by a large margin. On the other hand, to achieve the same level of accuracy, \ours spends a much smaller cost. That means, our methods requires much less (high-fidelity) simulations to generate the training examples. This is particularly useful for large-scale CFD applications, in which the simulation is known to be very expensive. 

To further confirm the gains in computational efficiency, we compared the cost of running the standard numerical methods for the topological optimization and CFD tasks. After active learning, our learned surrogate models give $42$x and $466$x speed-up in computing (predicting) the high-fidelity solution fields against the numerical methods. Even counting the whole active learning cost, with the growth of test cases, the average cost (time) of \ours in computing a solution quickly becomes much smaller than the standard numerical methods. See details in Sec. 2 of Appendix. 

\textbf{Non-Active Learning.}{ Finally, to confirm the capacity of our deep NN model,  we compared with state-of-the-art high-dimensional non-active learning surrogate models. Note that these models all lack effective active learning strategies. Our model consistently outperforms these methods, often by a large margin (Sec. 3 of the Appendix).}
\vspace{-0.1in}
\section{Conclusion}
\vspace{-0.1in}
We have presented \ours, a deep multi-fidelity active learning approach for high-dimensional outputs. Our deep neural network based multi-fidelity model is flexibly enough to capture the strong, complex relationships between the outputs and between the fidelities. We proposed a mutual information based acquisition function that accounts for multi-fidelity queries. To calculate and optimize the acquisition function, we developed an efficient and reliable method that successfully overcomes the computational challenges due to the massive outputs.

\bibliographystyle{apalike}
\bibliography{DeepMFAL}

\onecolumn
\aistatstitle{Appendix}

\section{Experimental Details}
\subsection{Solving Partial Differential Equations}

\noindent \textbf{Burgers' equation} is a canonical nonlinear hyperbolic PDE, widely used to model various physical phenomena, such as nonlinear acoustics~\citep{sugimoto1991burgers},  fluid dynamics~\citep{chung2010computational},  and traffic flows~\citep{nagel1996particle}. Due to its capability of developing discontinuities  (\ie shock waves), Burger's equation is used as a benchmark test example for many numerical solvers and surrogate models~\citep{kutluay1999numerical,shah2017reduced,raissi2017physics}.The viscous version of Burger's equation is given by 
\[
\frac{\partial u}{\partial t} + u \frac{\partial u}{\partial x} = v \frac{\partial^2 u}{\partial x^2},
\]
where $u$ is the volume, $x$ is a spatial location, $t$ is the time, and $v$ is the viscosity. We set $x\in[0,1]$, $t \in [0,3]$, and $u(x,0)=\sin(x\pi/2)$ with a homogeneous Dirichlet boundary condition.  The input parameter is the viscosity $v \in [0.001, 0.1]$. Given the input, we aim to predict the solution field (\ie the values of $u$) in the spatial-temporal domain $[0, 1] \times [0, 3]$. To obtain the training and test datasets, we solve the equation using the finite element~\citep{zienkiewicz1977finite} with hat functions in space and backward Euler in time domains on a regular mesh. 

\noindent \textbf{Poisson's equation} is an elliptic PDE and commonly used to model potential fields, \eg  electrostatic and gravitational fields, in physics and mechanical engineering~\citep{chapra2010numerical}. The equation used in our experiment is given by
\[
\Delta u = \beta \delta(\x - \c),
\]
where $\Delta $ is the Laplace operator~\citep{persides1973laplace}, $\u$ is the volume, $\delta(\cdot)$ is the Dirac-delta function, and $\c$ is the center of the domain. We  used a 2D spatial domain, $\x \in [0, 1] \times [0, 1$], and Dirichlet boundary conditions. We used  the constant values of the four boundaries and $\beta$ as the input parameters, each of which ranges from $0.1$ to $0.9$. We solved the equation using the finite difference method with the first order center differencing scheme and regular rectangle meshes.

\noindent \textbf{Heat equation} is a fundamental PDE that models heat conduction over time. It is also widely used in many other areas, such as probability theory~\citep{spitzer1964electrostatic,burdzy2004heat} and financial mathematics~\citep{black1973pricing}. 
The equation is defined as
\[
\frac{\partial u}{\partial t} + \alpha \Delta u =0,
\]  
where $u$ is the heat, $\alpha$ the thermal conductivity, and $\Delta$ the Laplace operator. In our experiment, we used  a 2D spatial-temporal domain $x\in[0,1]$, $t \in [0,5]$ with the Neumann boundary condition at $x=0$ and $x=1$, and $u(x,0)=H(x-0.25)-H(x-0.75)$, where $H(\cdot)$ is the Heaviside step function. We considered three input parameters ---  the flux rate $\in [0, 1]$ of the left boundary at $x=0$, the flux rate $\in [-1, 0]$ of the right boundary at $x=1$, and $\alpha \in [0,01, 0.1]$.
To generate the training and test examples, we solve the equation with the finite difference in the space domain and backward Euler in the time domain. 

\begin{figure*}[htbp!]
	\centering
	\includegraphics[width=1.0\textwidth]{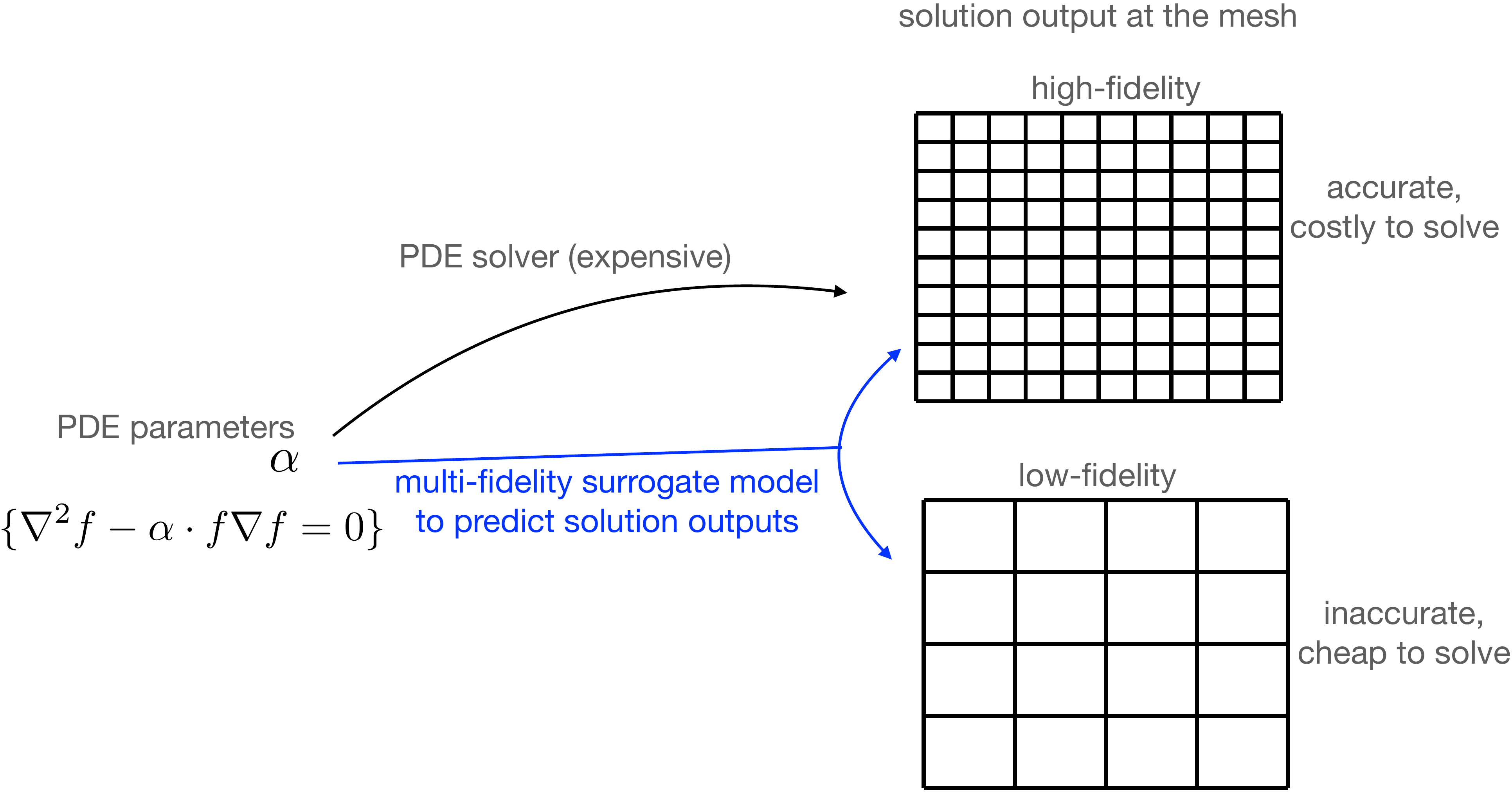}
	\caption{\small Illustration of the motivation and goal with physical simulation as an example. Solving every PDE from scratch is expensive. Hence, we  learn a multi-fidelity surrogate model that can predict high-fidelity solution outputs outright given the PDE parameters. We develop active learning methods to further reduce the cost of running numerical solvers to generate/collect training examples. }
	\label{fig:illustration}
\end{figure*}
\begin{figure*}[htbp!]
	\centering
	\includegraphics[width=0.6\textwidth]{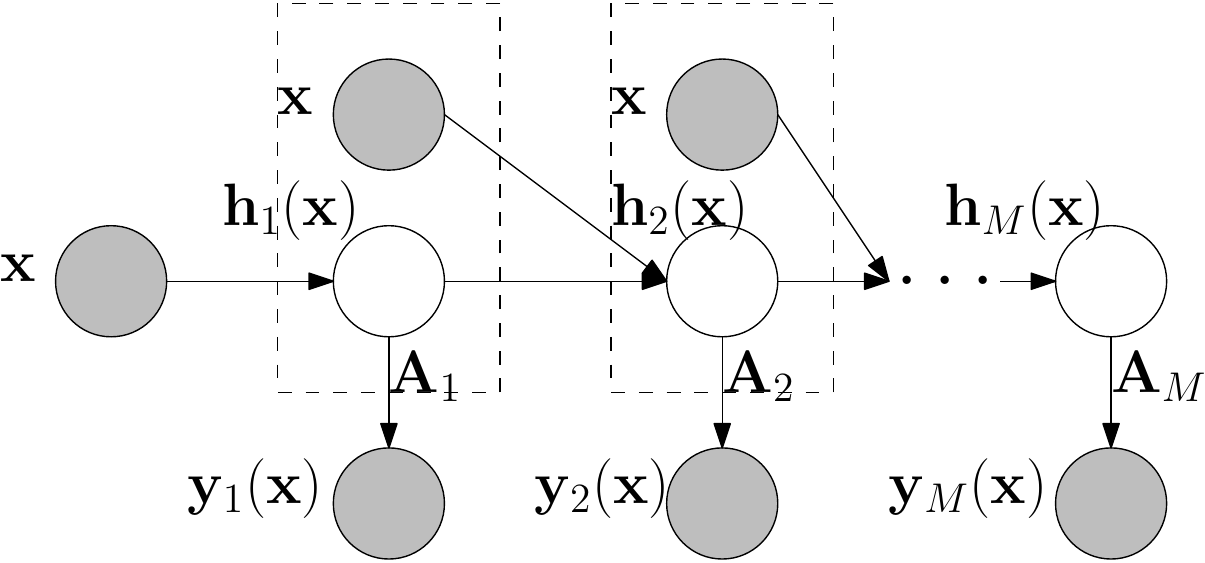}
	\caption{\small Graphical representation of the deep multi-fidelity model. The low dimensional latent output in each fidelity $\h_m(\x)$ ($1\le m \le M$) is generated by a (deep) neural network.} \label{fig:graphical}
\end{figure*}
\subsection{Predicting Fluid Dynamics}
\begin{figure*}
	\centering
	\setlength\tabcolsep{0pt}
	\begin{tabular}[c]{ccc}
		\setcounter{subfigure}{0}
		\begin{subfigure}[t]{0.33\textwidth}
			\centering
			\includegraphics[width=\textwidth]{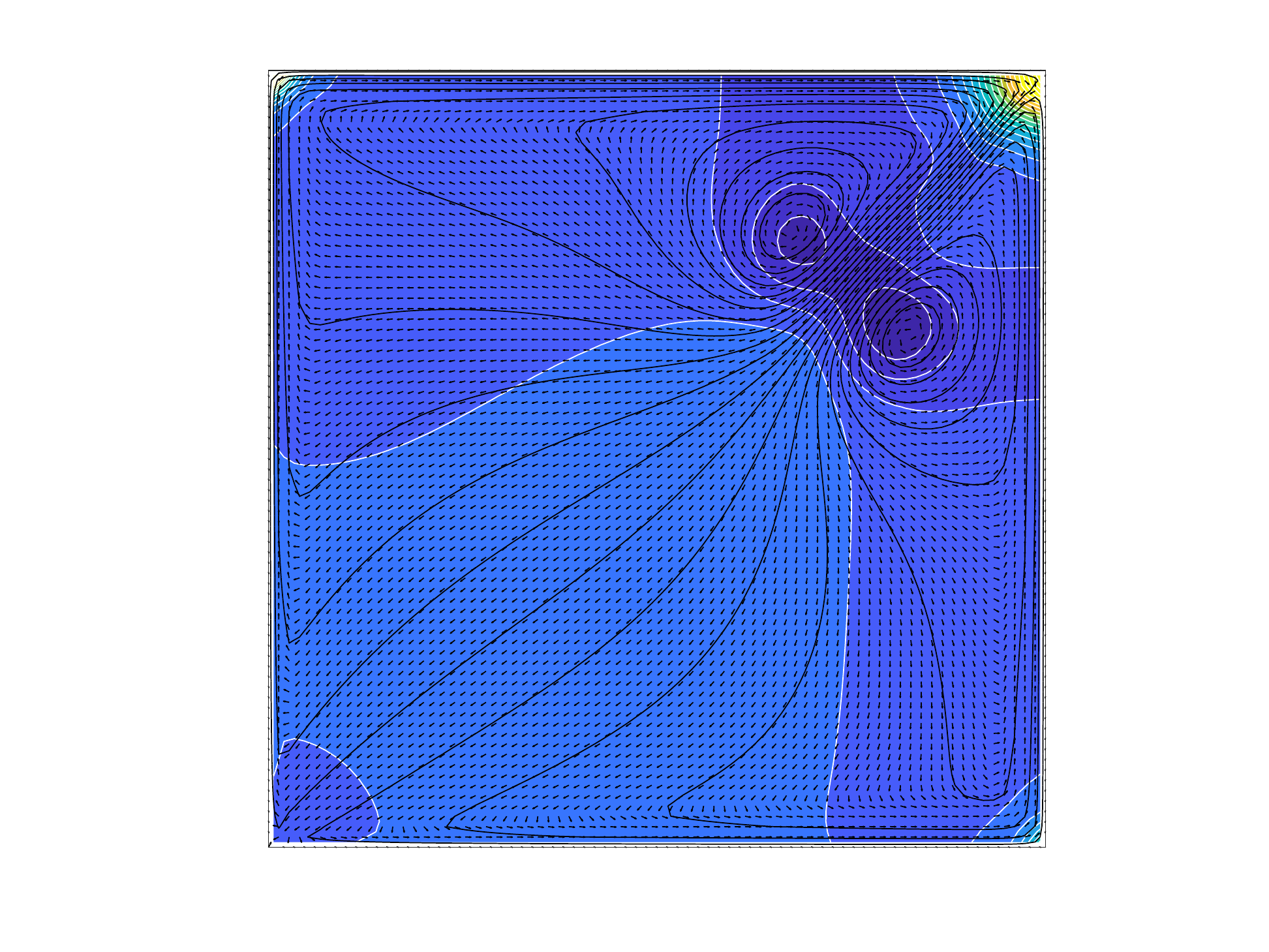}
			\caption{$t=1$}
		\end{subfigure} &
		\begin{subfigure}[t]{0.33\textwidth}
			\centering
			\includegraphics[width=\textwidth]{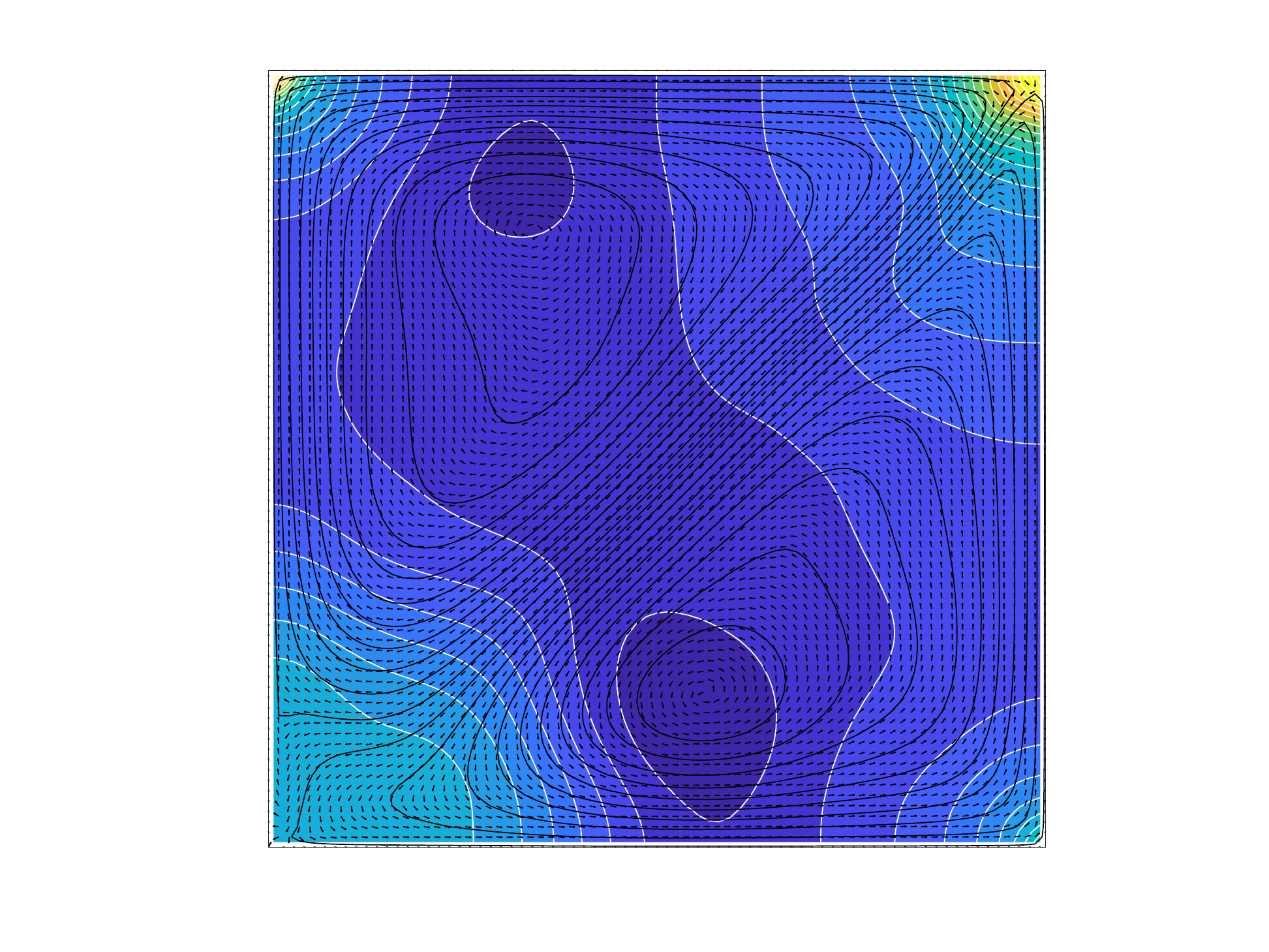}
			\caption{$t=5$}
		\end{subfigure}
		&
		\begin{subfigure}[t]{0.33\textwidth}
			\centering
			\includegraphics[width=\textwidth]{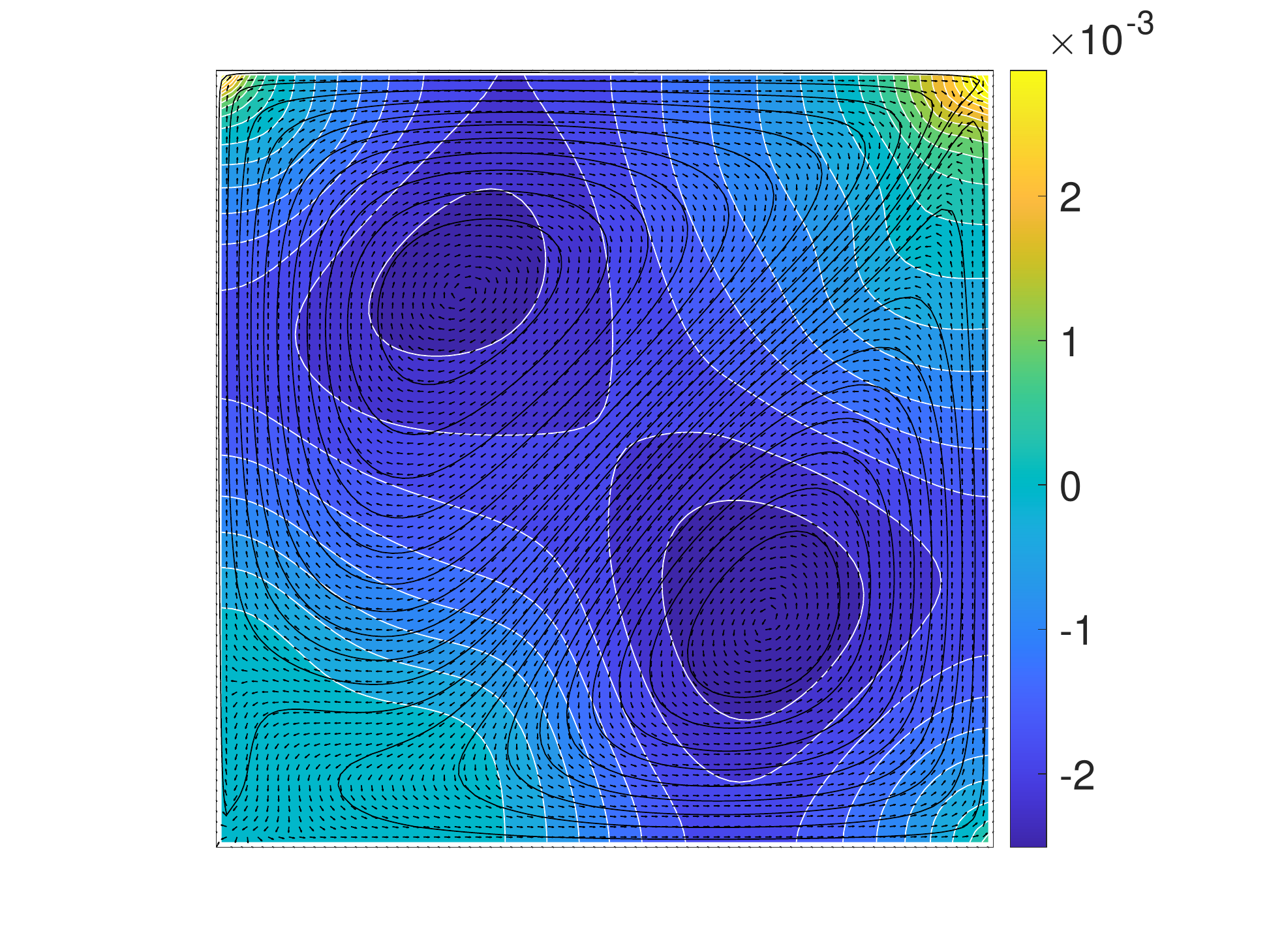}
			\caption{$t=9$}
		\end{subfigure}
	\end{tabular}
	\caption{\small Examples of the first component of the velocity field (with contour lines) at three time points.} \label{fig:cfd-example}
\end{figure*}

We also examined \ours in predicting the velocity field of a flow within a rectangular domain with a prescribed velocity along the boundaries~\citep{bozeman1973numerical}. This is a classical computational fluid dynamics (CFD) problem. The simulation of the flow involves solving the incompressible Navier-Stokes (NS) equation~\citep{chorin1968numerical}, 
\[
\rho (\textbf{u} \cdot \nabla) \textbf{u} = -\nabla p + \mu \nabla^2 \textbf{u},
\]
where $\rho$ is the density, $p$ is the pressure, $\textbf{u}$ is the velocity, and $\mu$ is the dynamic viscosity. The equation is well known to be challenging to solve due to their complicated behaviours under large Reynolds numbers.
We set the rectangular domain to $[0, 1] \times [0, 1]$, and time $t \in [0, 10]$. 
The input includes the tangential velocities of the four boundaries and the Reynold number $\in [100, 5000]$. The output are the first component of the velocity field at $20$ equally spaced time steps in $[0, 10]$ (see Fig. \ref{fig:cfd-example}). To generate the training and test examples, we used the SIMPLE algorithm~\citep{caretto1973two} with a stagger grid~\citep{versteeg2007introduction}, the up-wind scheme~\citep{versteeg2007introduction} for the spatial difference, and the implicit time scheme with fixed time steps to solve the NS equation. 
\begin{figure*}
	\centering
	\setlength\tabcolsep{0pt}
	\begin{tabular}[c]{ccc}
		\setcounter{subfigure}{0}
		\begin{subfigure}[t]{0.38\textwidth}
			\centering
			\includegraphics[width=\textwidth]{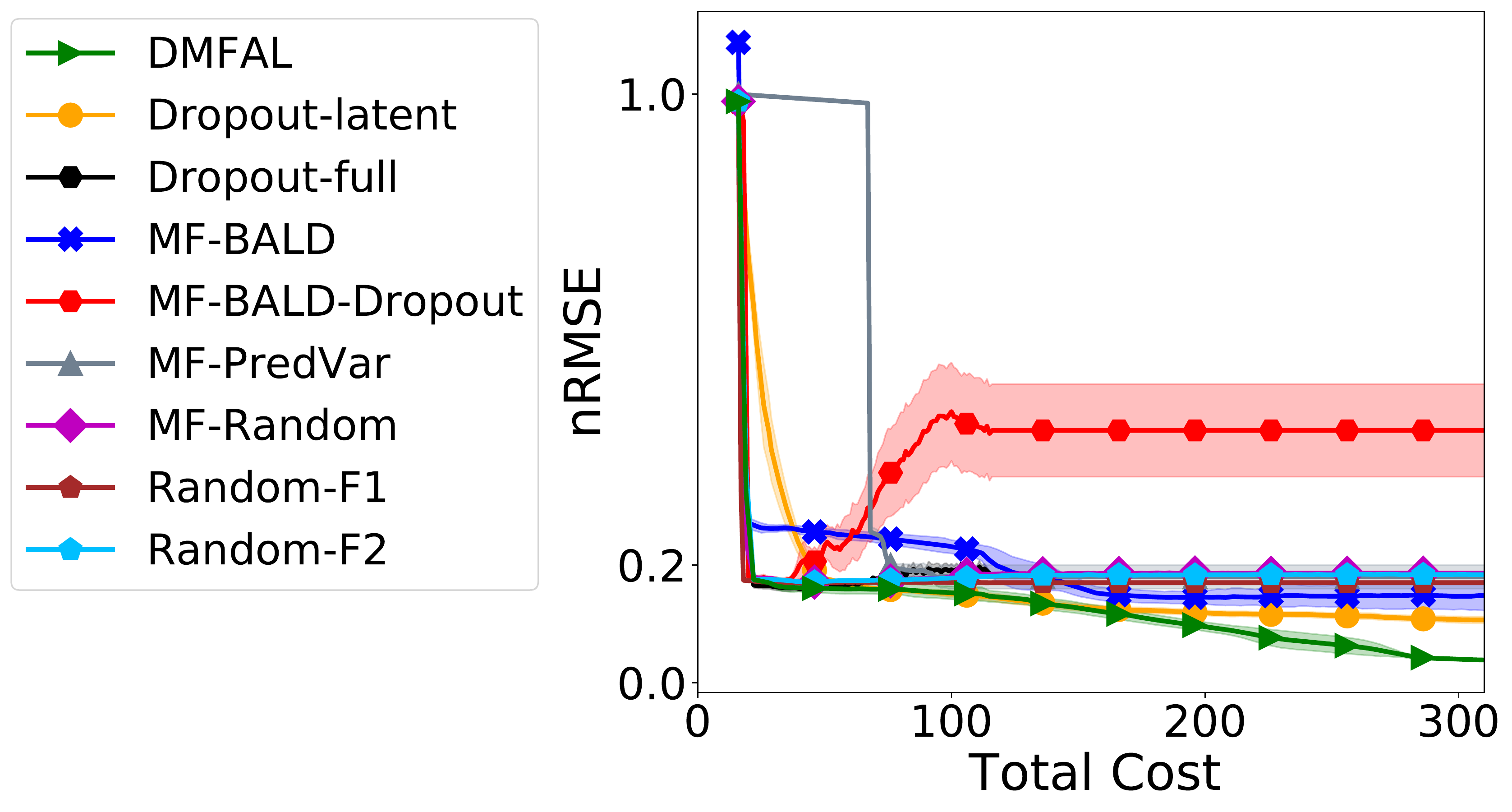}
			\caption{Burgers' equation}
		\end{subfigure} &
		\begin{subfigure}[t]{0.25\textwidth}
			\centering
			\includegraphics[width=\textwidth]{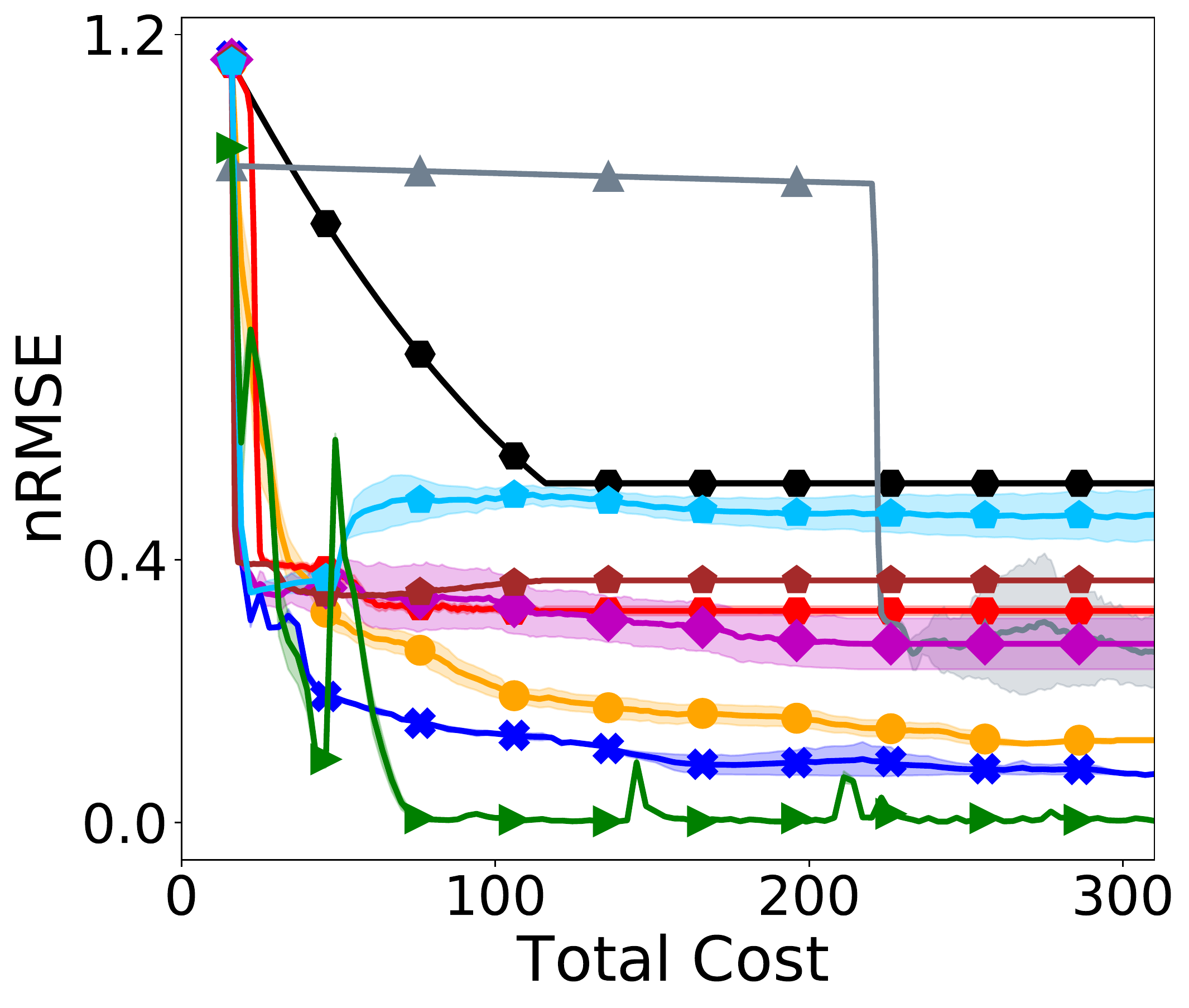}
			\caption{Heat equation}
		\end{subfigure}
		&
		\begin{subfigure}[t]{0.25\textwidth}
			\centering
			\includegraphics[width=\textwidth]{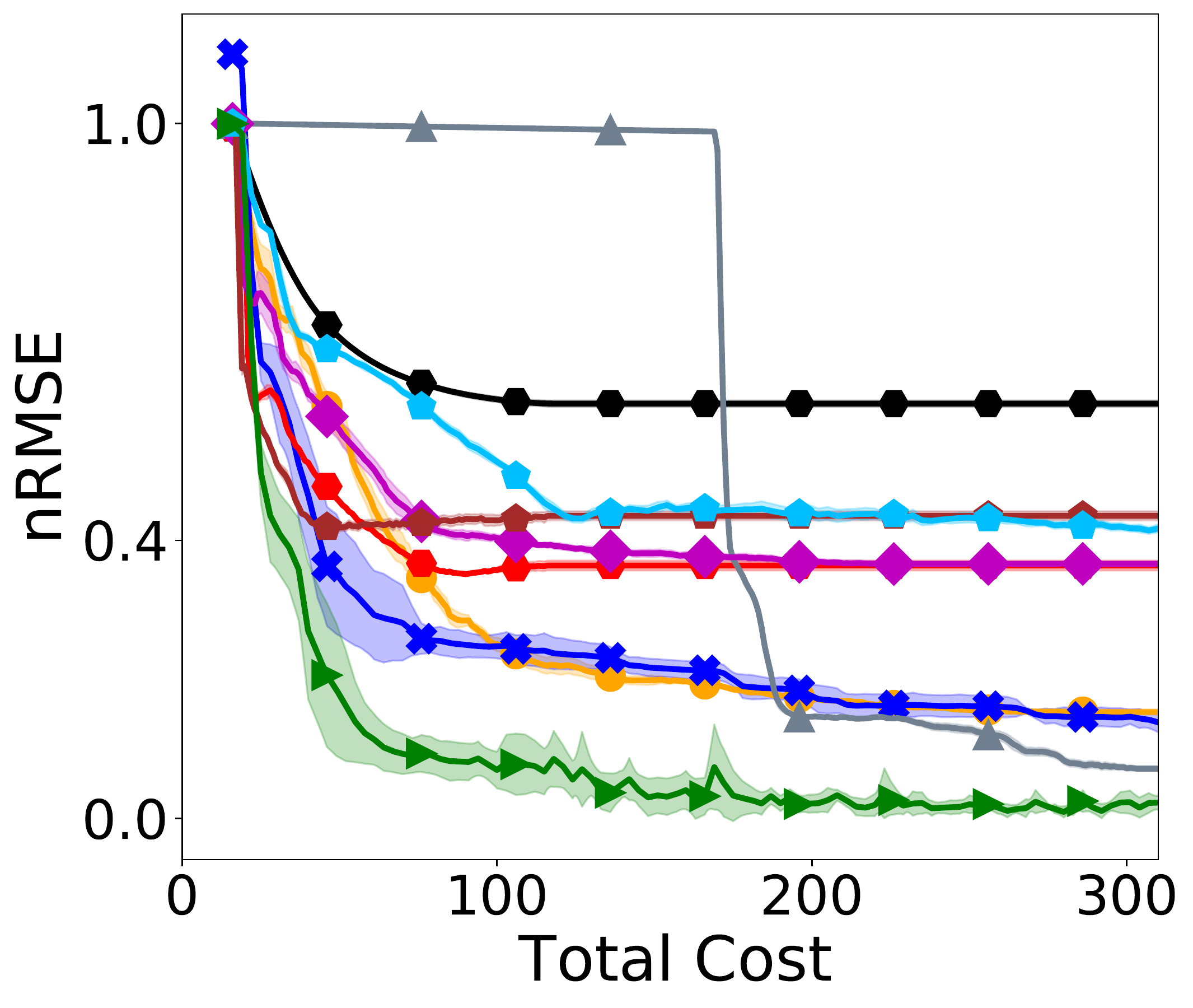}
			\caption{Poisson's equation}
		\end{subfigure}
	\end{tabular}
	\caption{\small Normalized root-mean-square error (nRMSE) for active learning of PDE solution fields with two-fidelity queries. The normalizer is the mean of the test outputs. The results are averaged from five runs. The shaded regions indicate the standard deviations.} \label{fig:solving-pde-2fid-all}
	\vspace{-0.05in}
\end{figure*}
\subsection{Using Full Dropout}
\begin{figure}
	\centering
	\includegraphics[width=0.48\textwidth]{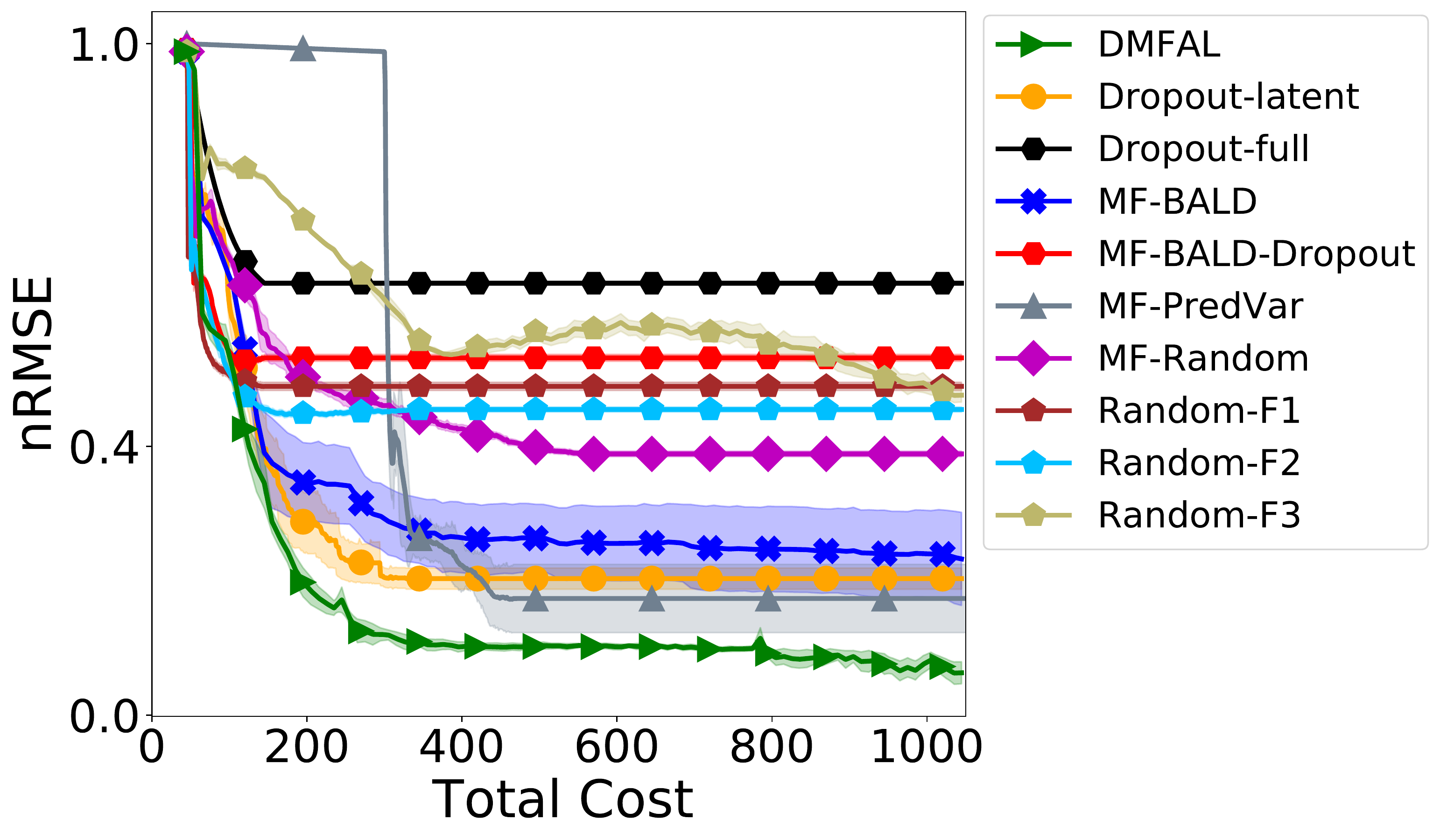}
	\caption{\small Normalized root-mean-square error (nRMSE) for active learning of the solution field of Poisson's equation, with three-fidelity queries. The results are averaged from five runs. }
	\label{fig:solving-po-3fid-all}
\end{figure}

In the experiment of Section 6.1 of the main paper, we have also applied MC dropout to directly generate the posterior samples of the final output in each fidelity $\{\y_m\}_{m=1}^M$. We generated $100$ samples.  We used these high-dimensional samples to calculate the empirical  means and covariance matrices , based on which we estimated a multi-variate Gaussian posterior for each $\y_m$ and $\widehat{\y}_m = [\y_m, \y_M]$ via moment matching.
These posterior distributions are then used to calculate and optimize our acquisition function and multi-fidelity BALD in the active learning. We denote these methods by {Dropout-full} and {MF-BALD-Dropout}. We report their nRMSE \textit{vs.} cost in Fig. \ref{fig:solving-pde-2fid-all} and \ref{fig:solving-po-3fid-all}, along with all the other methods. As we can see, in all the cases, {Dropout-full} is far worse than {Dropout-latent} and even inferior to random query strategies, except that in Fig. \ref{fig:solving-pde-2fid-all}a, they are quite close.
 The prediction accuracy of {MF-BALD}  is always much better than {MF-BALD-Dropout}. Those results demonstrate that the posterior distributions of the high-dimensional outputs, if fully estimated by a small number of dropout samples, are far less accurate and reliable than our method. Also, the computation is much more costly --- we need to directly compute an empirical covariance matrix based on these high-dimensional samples, and calculate the log determinant. 

\section{Gains in Computational Efficiency}
\begin{figure*}
	\centering
	\setlength\tabcolsep{0pt}
	\begin{tabular}[c]{ccc}
		\setcounter{subfigure}{0}
		\begin{subfigure}[t]{0.333\textwidth}
			\centering
			\includegraphics[width=\textwidth]{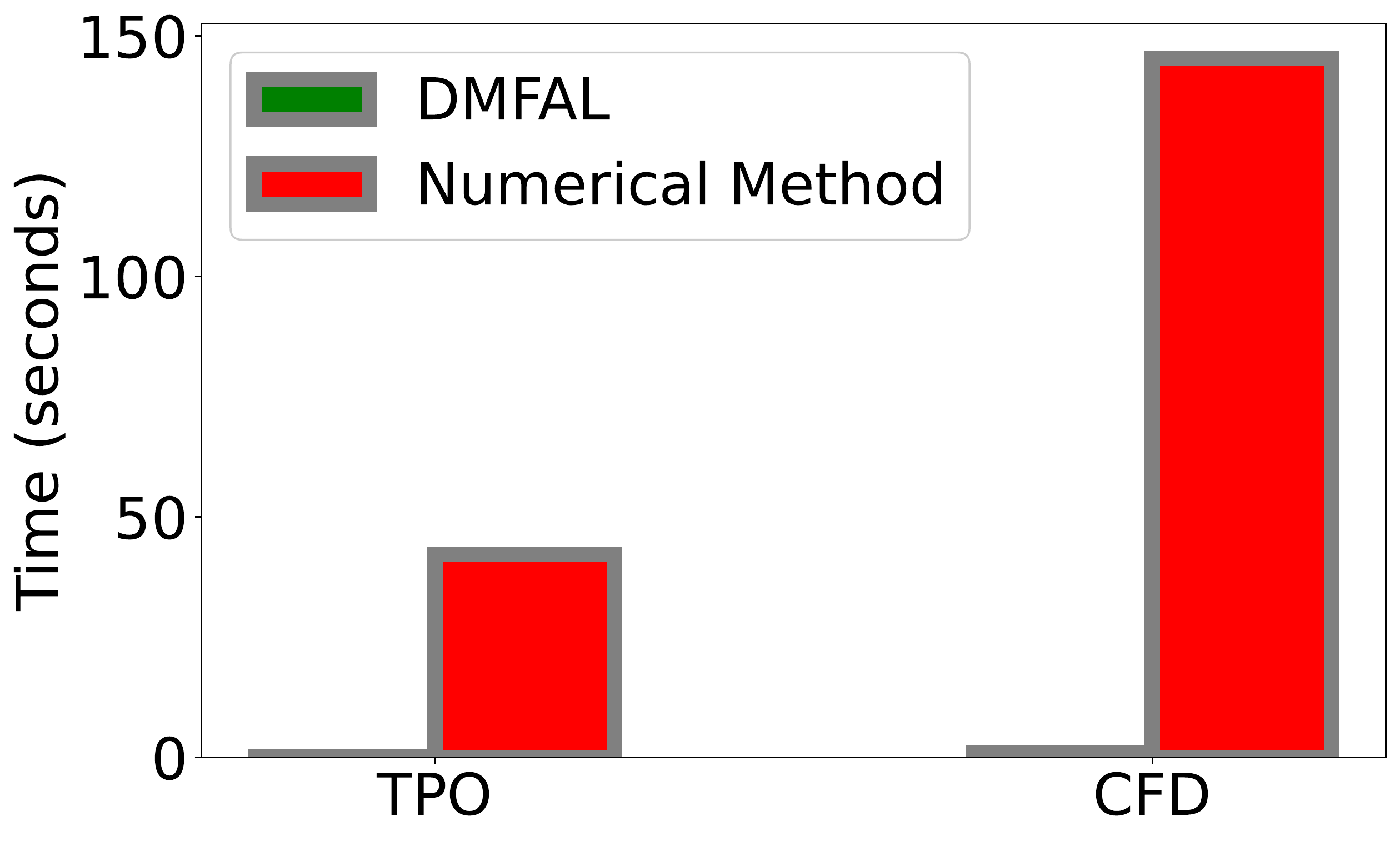}
			\caption{Speed of computing solutions}
		\end{subfigure} &
		\begin{subfigure}[t]{0.333\textwidth}
			\centering
			\includegraphics[width=\textwidth]{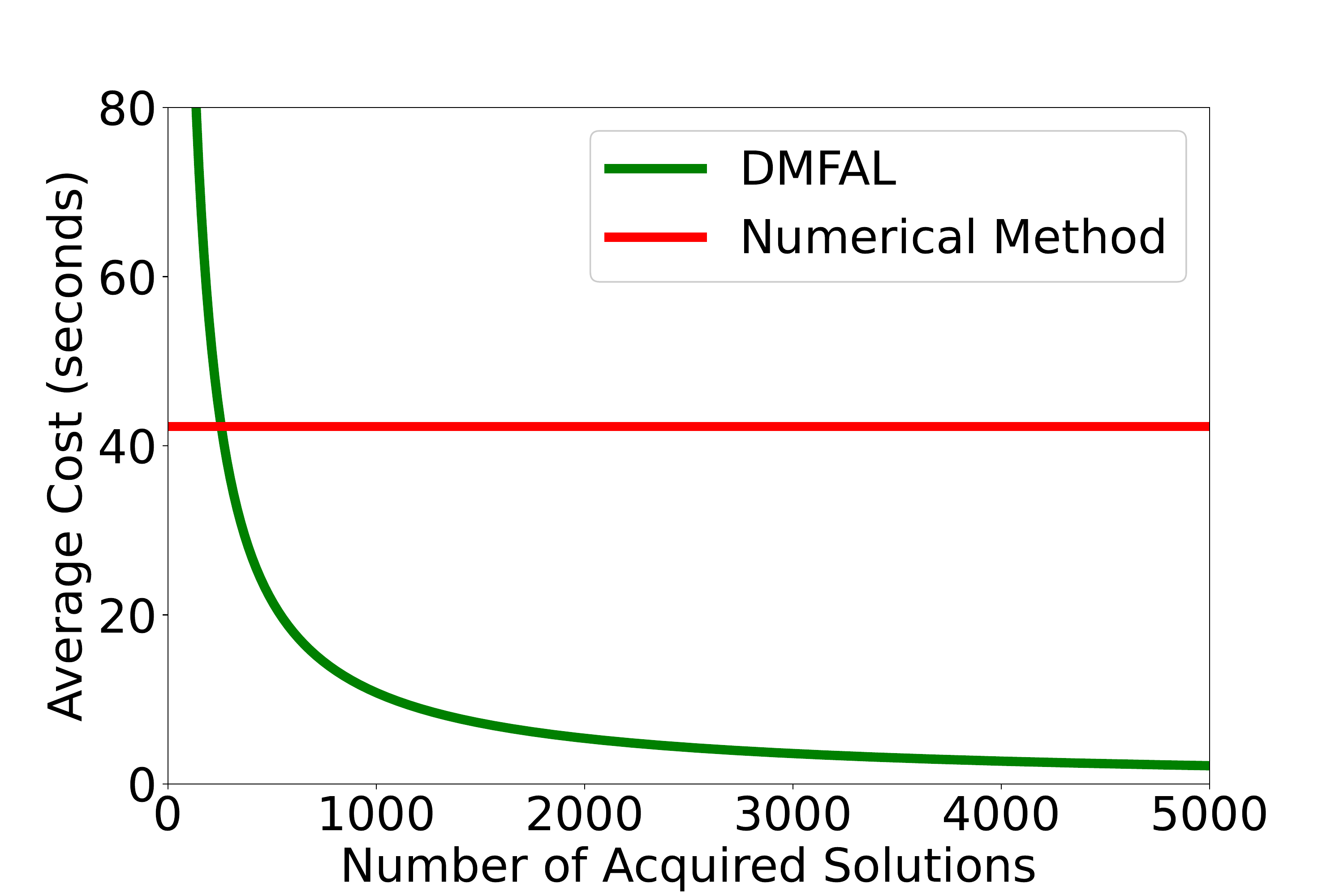}
			\caption{Average cost in \textit{topology structure optimization}}
		\end{subfigure} &
		\begin{subfigure}[t]{0.333\textwidth}
			\centering
			\includegraphics[width=\textwidth]{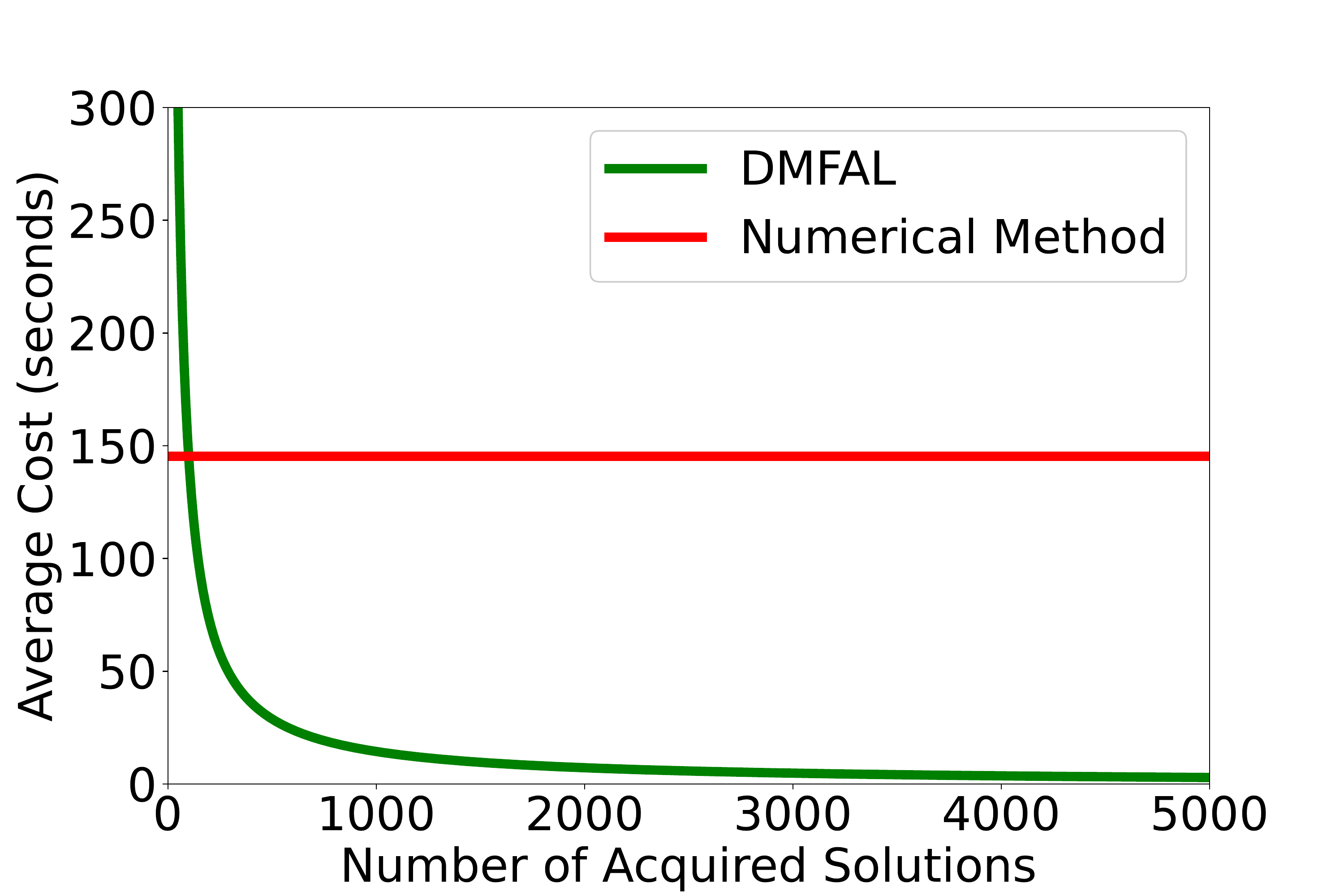}
			\caption{Average cost in \textit{Fluid dynamics}}
		\end{subfigure}
	\end{tabular}
	\caption{\small Speed of computing solutions after active learning (a) and total average solving (or predicting) cost with active learning included.} \label{fig:cost-ratio}
\end{figure*}

To confirm the gains in computational efficiency, we compared the cost of running standard numerical methods for topology structure optimization and fluid dynamics (Sec 6.2 and 6.3 in the main paper).  First,  after the active learning, we examined the average time of our learned model in computing (or predicting) a high-fidelity solution field, and contrast to that of the standard numerical approaches. As shown in Fig. \ref{fig:cost-ratio}a,   our model is mush faster, giving 42x and 466x speed-up for the two tasks.   Next, we consider adding the cost (running time) of the active learning, and calculate the average cost of predicting a high-fidelity solution field, $$\text{Cost}_{\text{avg}} = \frac{\text{Active Learning Cost} + \text{Prediction Cost}}{\text{Number of Acquired Solutions}}.$$ Note that for the numerical methods, there is no learning procedures and the active learning cost is zero. We examined how the average cost varies along with more acquired solutions (\ie test cases). As we can see in Fig. \ref{fig:cost-ratio} b and c, when only a few solutions are needed, directly running numerical methods is more economic  because we do not need to conduct an extra learning procedure. However, when we need to compute more and more solutions (test cases), the average cost of using the surrogate model quickly becomes much smaller, and keeps decreasing. That is because after training,  the surrogate model can predict solutions way more efficiently than running numerical methods from scratch, and the training cost will be diluted by the number of solutions acquired,  finally becoming negligible.  All these results have demonstrated the gains of computational efficiency of our method. 

\section{Non-Active Learning}
To confirm the surrogate performance of our deep NN model, we also compared with several state-of-the-art high-dimensional non-active learning surrogate models: (1) PCA-GP~\citep{higdon2008computer}, (2) IsoMap-GP~\citep{xing2015reduced}, (3) KPCA-GP~\citep{xing2016manifold}, and (4) HOGP~\citep{zhe2019scalable}. PCA-GP, IsoMap-GP and KPCA-GP are based on the classical linear model of of coregionalization (LMC)~\citep{journel1978mining}, and obtain the bases or low-rank structures from Principal Component Analysis (PCA),  IsoMap~\citep{balasubramanian2002isomap} and Kernel PCA~\citep{scholkopf1998nonlinear}, respectively. HOGP tensorizes the outputs, generalizes a multilinear Bayesian model with the kernel trick,  and  is flexible enough to capture nonlinear output correlations and can efficiently deal with very high-dimensional outputs. These methods are all single-fidelity models. In addition, we also compare with  (6) MFHoGP~\citep{wang2021multi}, a state-of-the-art multi-fidelity high-dimensional surrogate model based on matrix GPs. Note that MFHoGP does not support varying output dimensions across the fidelities. 

\begin{figure*}
	\centering
	\setlength\tabcolsep{0pt}
	\includegraphics[width=0.7\textwidth]{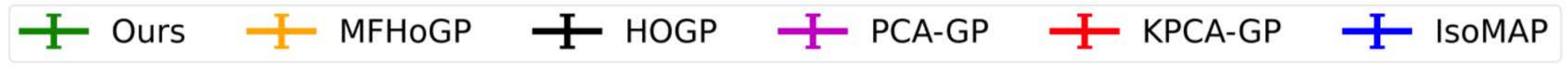}
	\begin{tabular}[c]{ccc}
		\setcounter{subfigure}{0}
		\begin{subfigure}[t]{0.33\textwidth}
			\centering
			\includegraphics[width=\textwidth]{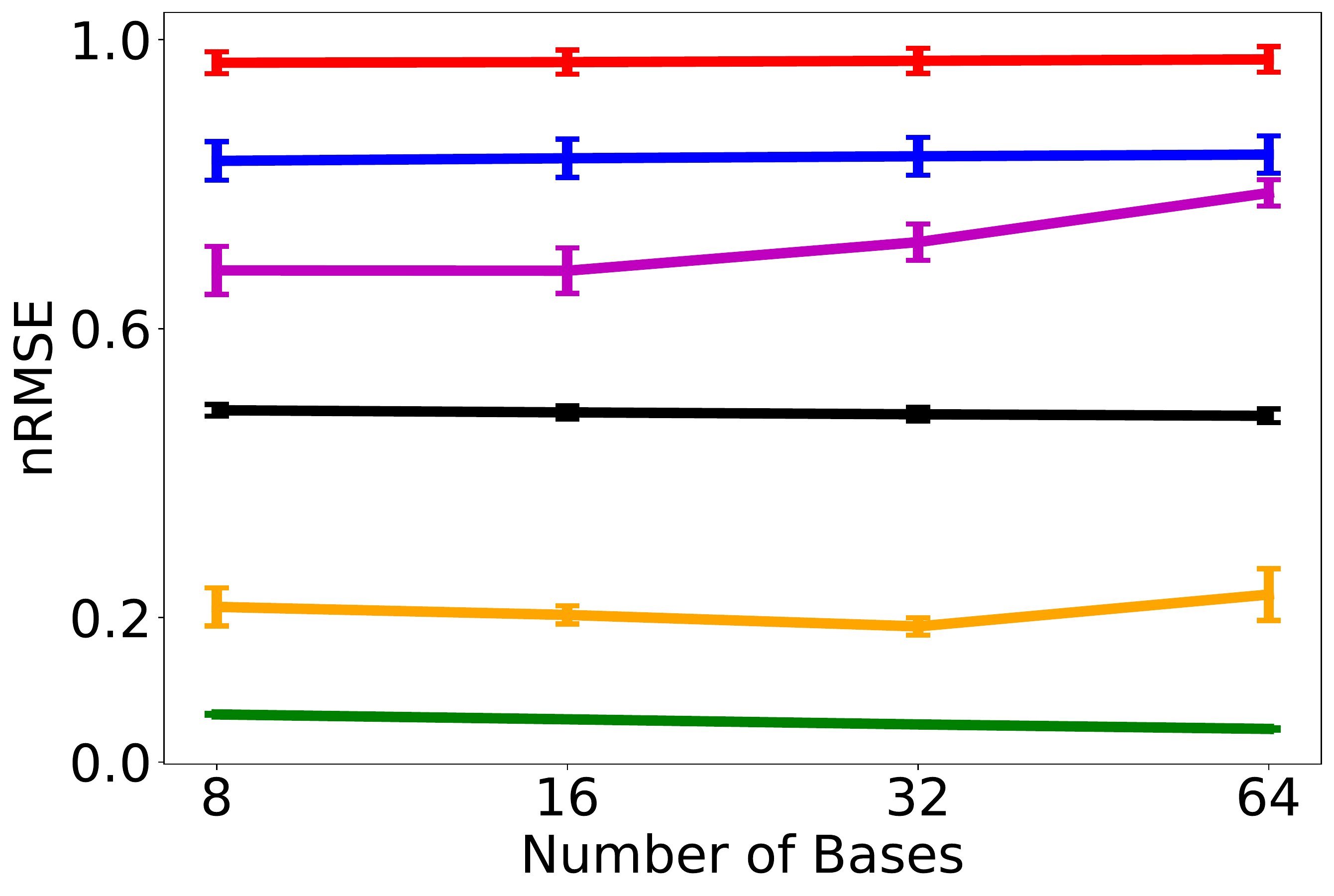}
			\caption{\textit{Poisson-3}}
		\end{subfigure}
		&
		\begin{subfigure}[t]{0.33\textwidth}
			\centering
			\includegraphics[width=\textwidth]{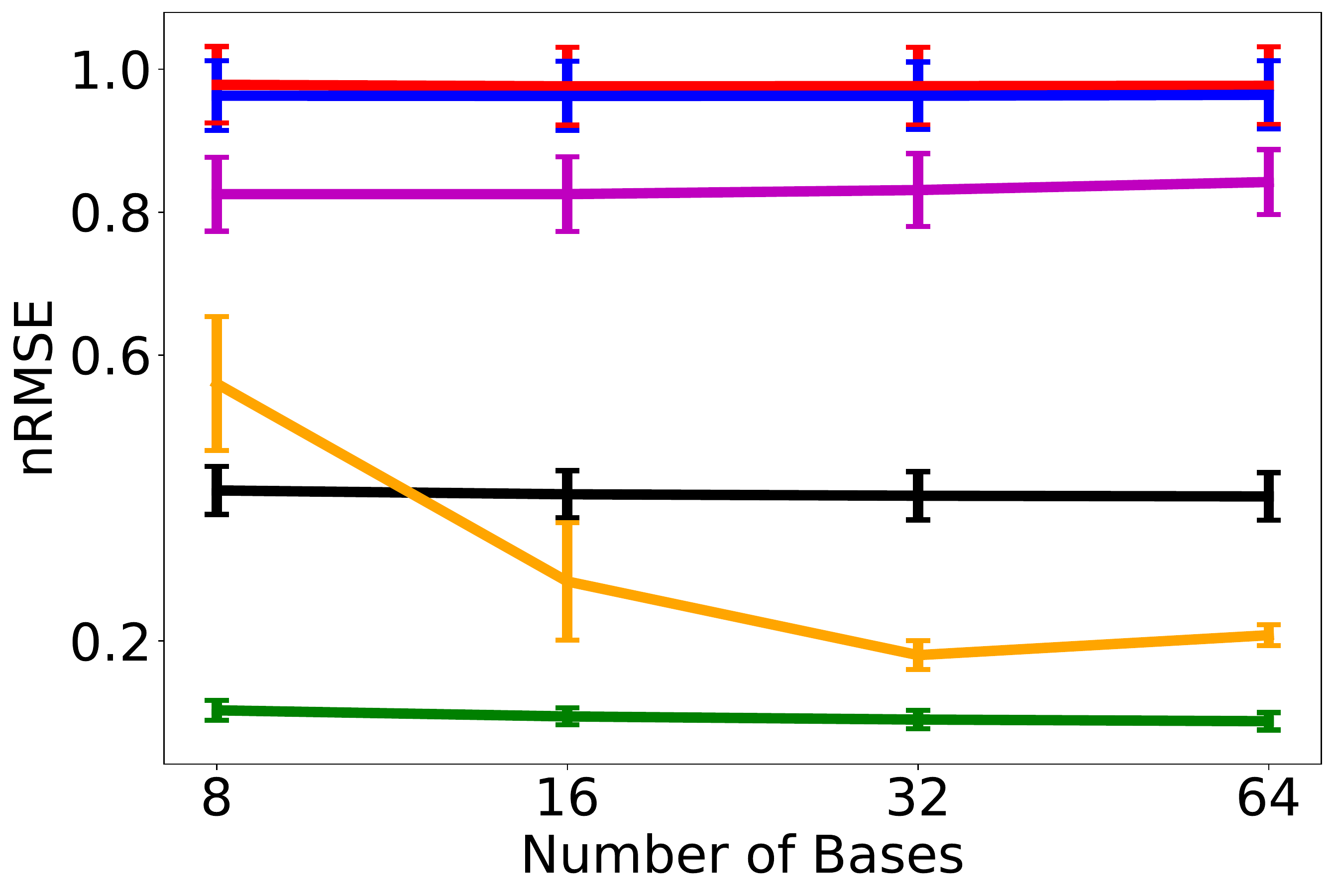}
			\caption{\textit{Topology structure optimization}}
		\end{subfigure} 
		&
		\begin{subfigure}[t]{0.33\textwidth}
			\centering
			\includegraphics[width=\textwidth]{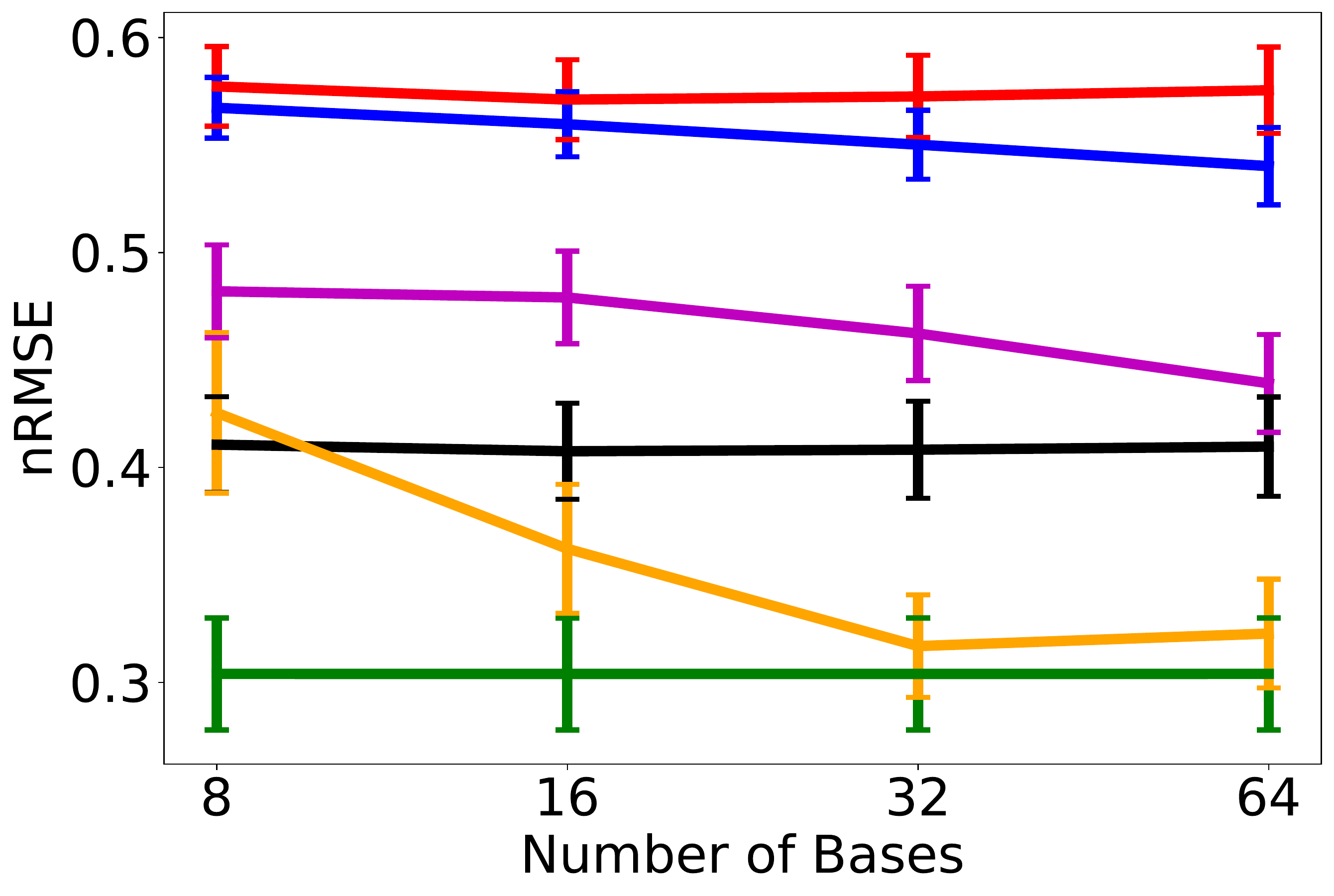}
			\caption{\textit{Fluid dynamics}}
		\end{subfigure}
	\end{tabular}
	\caption{\small Predictive performance of non-active learning. The results were averaged over five runs.} \label{fig:na-predict2}
\end{figure*}

We tested all these methods in the three applications: \textit{Poisson-3}, \textit{Topology structure optimization} and \textit{Fluid dynamics} (see Sec. 6.1, 6.2 and 6.3 of the main paper). For \textit{Poisson-3}, we randomly generated 43, 43, 42 examples for the first, second and third fidelity, respectively.  The inputs were uniformly sampled from the input space. For \textit{Topology structure optimization} and \textit{Fluid dynamics}, we randomly generated 64 examples for each fidelity (two fidelities in total). For each application, we randomly generated 512 examples for testing. The settings of those fidelities are the same as in the main paper. Since MFHoGP does not support varying output dimensions in different fidelities, we interpolated the training outputs at low fidelities to ensure all the fidelities have the same output dimension. We varied the number of bases (required by all the competing methods) from \{8, 16, 32, 64\}, which corresponds to the dimension of the latent output in our model (see Sec. 3 of the main paper). We ran the experiments for five times, and report the average nRMSE and its standard deviation in Fig. \ref{fig:na-predict2}. As we can see, our model consistently outperforms all the methods, often by a large margin. MFHoGP is the second best, but in most cases, its prediction accuracy is still significantly worse than our model ($p<0.05$). The remaining single-fidelity models are almost always worse than the multi-fidelity models --- that is reasonable, because the former cannot differentiate examples of different fidelities, blindly mix them in the training, and hence cannot leverage their relationships to improve the prediction. Together these results have shown the advantage of our deep surrogate model, even in the non-active learning setting. 

\section{Ablation Study about Multi-Variate Delta Method}
We examined the rationality of the multi-variate delta method used in our work (see (6) in Section 4.2 of the main paper). To this end, we used a Bayesian neural networks (BNN) to learn two nontrivial benchmark functions, \textit{Branin} and \textit{Levy}, which are defined as follows.  For \textit{Branin}, the input $\x \in [-5, 10]\times [0, 15]$, and for \textit{Levy}, $\x \in [-10, 10]^2$. We used three hidden layers, with 40 neurons per layer, and \texttt{tanh} as the activation function (the same as in our experiment).
\begin{align}
	y_{\text{Branin}}(\x) = -\left(\frac{-1.275x_1^2}{\pi^2} + \frac{5x_1}{\pi} + x_2 - 6\right)^2  - \left(10 - \frac{5}{4\pi}\right)\cos(x_1) - 10, \notag \\
	y_{\text{Levy}}(\x) = -\sin^2(3\pi x_1) - (x_1-1)^2[1 + \sin^2(3\pi x_2)] - (x_2 - 1)^2[1 + \sin^2(2\pi x_2).
\end{align}

\begin{figure*}
	\centering
	\setlength\tabcolsep{0pt}
	\begin{tabular}[c]{cc}
		\setcounter{subfigure}{0}
		\begin{subfigure}[t]{0.33\textwidth}
			\centering
			\includegraphics[width=\textwidth]{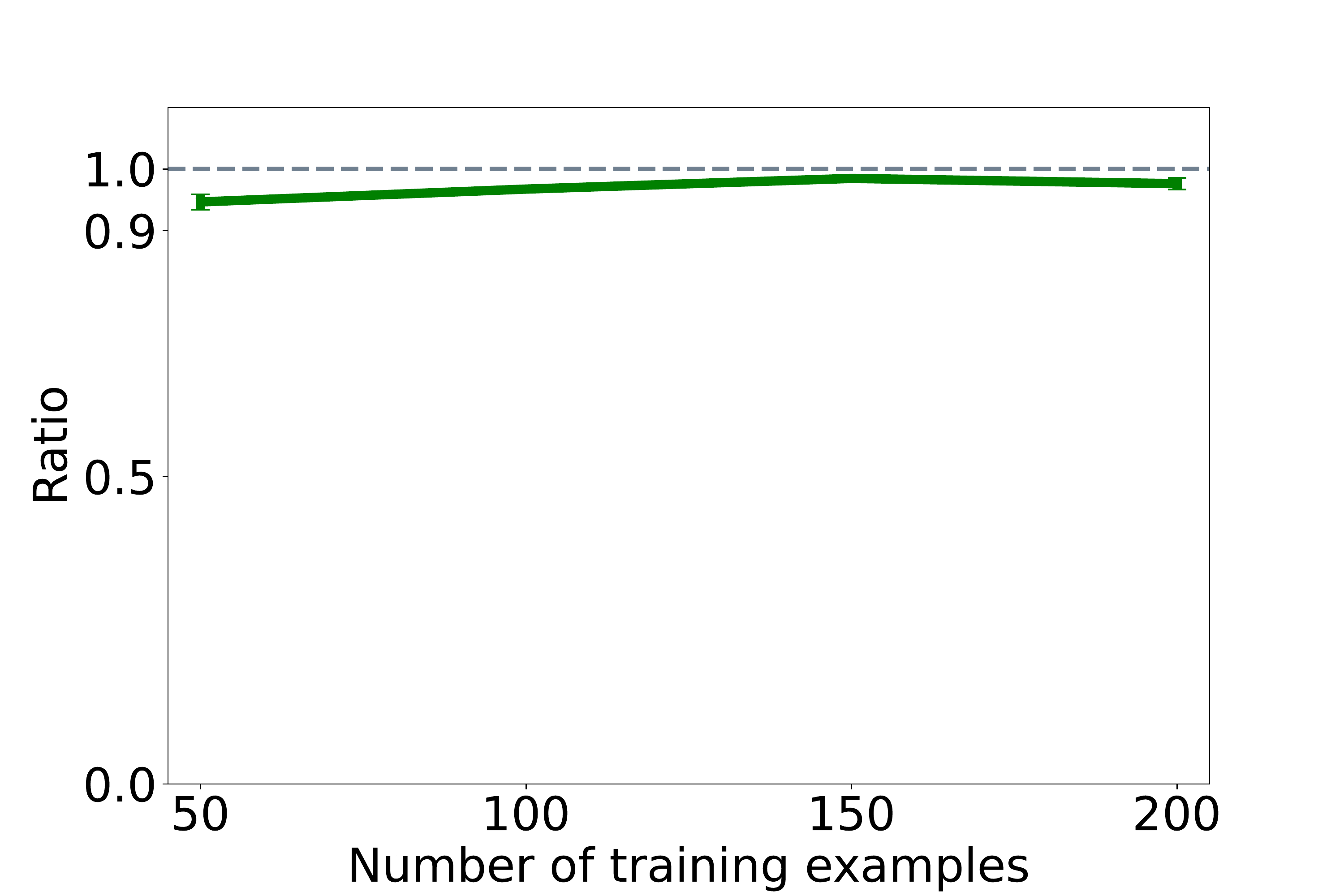}
			\caption{\textit{Branin}}
		\end{subfigure}
		&
		\begin{subfigure}[t]{0.33\textwidth}
			\centering
			\includegraphics[width=\textwidth]{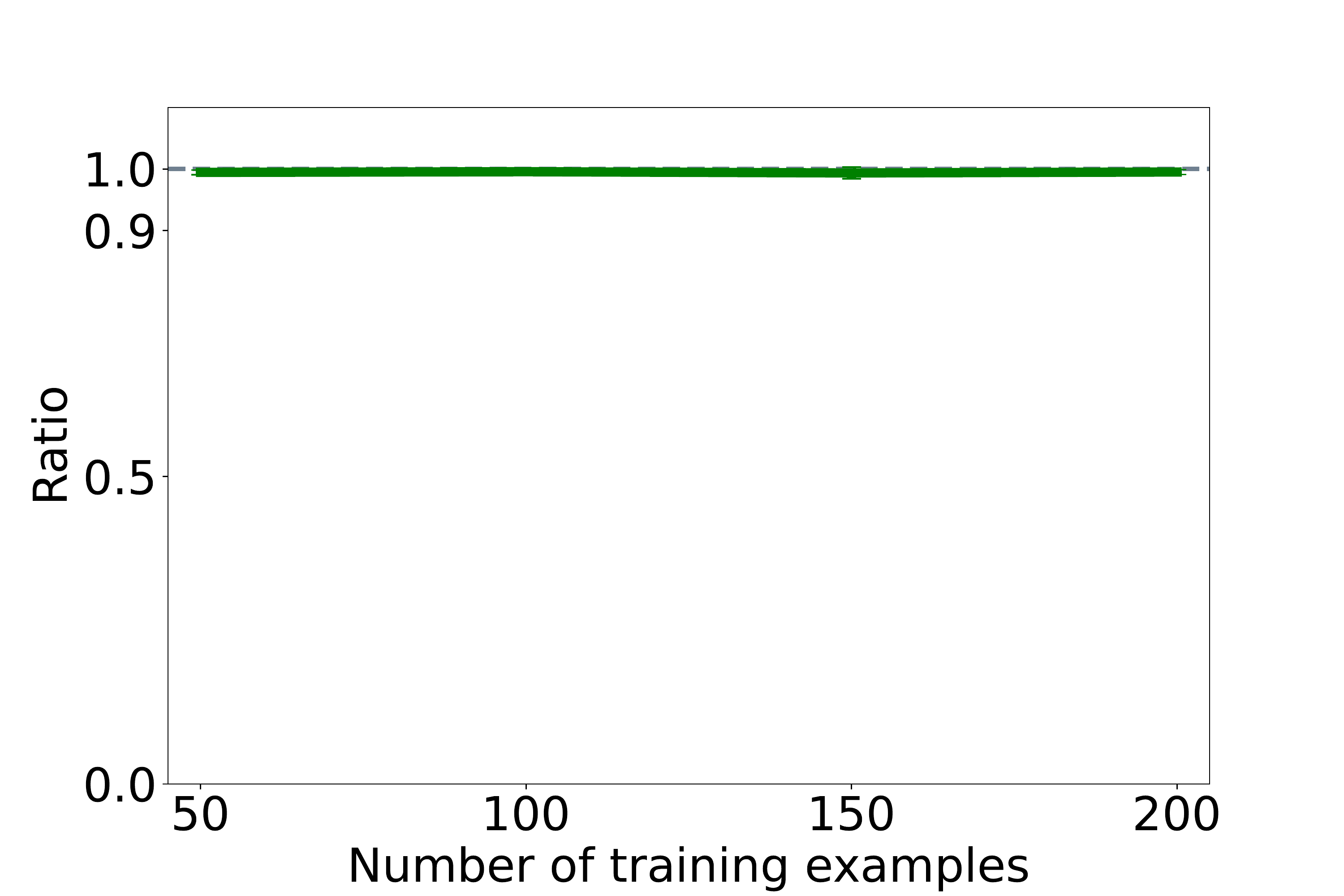}
			\caption{\textit{Levy}}
		\end{subfigure} 
	\end{tabular}
	\caption{\small The ratio between first-order and second-order approximations. \cmt{\textcolor{blue}{Shibo: change y-label to ``Ratio'', add ``0.9'' in the y-axis; x-label ``Number of training examples''}}} \label{fig:approx-ratio}
\end{figure*}

We generated $N$ training examples by randomly sampling from the input domain, and then used the structural variational inference to estimate the BNN. In our work,  we used the first-order Taylor expansion of the NN output to approximate the posterior mean and (co-)variance (see (6-8) in the main paper). To confirm the rationality, we examined the ratio between the first-order Taylor expansion and the second-order Taylor expansion. Specifically, given a new input $\x$, we first sample an instance of the NN weights $\Wcal$ from the learned posterior $p(\Wcal|\Dcal)$. Then, we expand the NN output $f_{\Wcal}(\x)$ at the posterior mean of the weights $\EE[\Wcal]$, and calculate 
\[
\text{Ratio} = \frac{|\text{First-Order Taylor Expansion of }  f_{\Wcal}(\x)|}{|\text{Second-Order Taylor Expansion of } f_{\Wcal}(\x) |}.
\]
We varied $N$ from $\{50, 100, 150, 200\}$. To obtain a reliable estimate of the ratio, for each $N$, we randomly sampled $100$ inputs. Given each input, we randomly drew $10$ samples of $\Wcal$ to calculate the ratio. We report the average ratio, and its standard deviation in Fig. \ref{fig:approx-ratio}. As we can see, in both cases, the ratio is constantly close to one, and the standard deviation is close to zero. That means, the first-order approximation dominates, and the second-order terms can be ignored. This is consistent with our analysis in the main paper.

%

\end{document}